\documentclass[journal]{IEEEtran}

\IEEEoverridecommandlockouts                              

\usepackage{amsmath}    
\usepackage{amsfonts}
\usepackage{graphicx}   
\usepackage{subcaption}
\usepackage{epsfig}
\usepackage{algorithmic}
\usepackage{color}
\usepackage[normalem]{ulem}
\usepackage{cancel}
\usepackage{amssymb}
\usepackage{color}

\usepackage[ruled,vlined,titlenotnumbered]{algorithm2e}
\usepackage{cite}
\usepackage{float}

\newcommand{\R}{\mathbb{R}}

\newcommand{\pcset}{\mathcal{U}_p} 
\newcommand{\tcset}{\mathcal{U}_s} 
\newcommand{\tcfset}{\mathbb{U}_s} 
\newcommand{\dset}{\mathcal{D}}
\newcommand{\dfset}{\mathbb{D}}

\newcommand{\pset}{\mathcal{P}} 
\newcommand{\tset}{\mathcal{S}} 

\newcommand{\tvar}{t}
\newcommand{\thor}{T} 

\newcommand{\astate}{\eta}
\newcommand{\estate}{e}
\newcommand{\tstate}{s} 
\newcommand{\pstate}{p} 
\newcommand{\rstate}{r} 
\newcommand{\rtrans}{\Phi}

\newcommand{\ttraj}{\xi_{\tdyn}} 
\newcommand{\rtraj}{\xi_\rdyn}

\newcommand{\senseDist}{M}

\newcommand{\tctrl}{u_s} 
\newcommand{\dstb}{d} 
\newcommand{\pctrl}{u_p} 

\newcommand{\tdyn}{f} 
\newcommand{\pdyn}{h} 
\newcommand{\rdyn}{g} 

\newcommand{\plannerfunc}{\text{nextState}}

\newcommand{\ptmat}{Q} 

\newcommand{\errfunc}{l} 
\newcommand{\valfunc}{V} 

\newcommand{\deriv}{\nabla\valfunc} 

\newcommand{\dt}{\Delta t} 

\newcommand{\tgoal}{\mathcal G}
\newcommand{\goal}{\tgoal_\pstate}
\newcommand{\contrgoal}{\tgoal_{\pstate,\text{contr}}}

\newcommand{\tconstr}{\mathcal C}
\newcommand{\constr}{\tconstr_\pstate}
\newcommand{\constrSense}{\tconstr_{\pstate,\text{sense}}}
\newcommand{\constrAug}{\tconstr_{\pstate,\text{aug}}}

\newcommand{\example}[1]%
{
	\textbf{Running example:}
	\textit{#1}
}

\newcommand{\TEB}{\mathcal B} 

\newtheorem{prop}{Proposition}

\title{\large \bf FaSTrack: a Modular Framework for Real-Time Motion Planning and Guaranteed Safe Tracking}

\author{Mo Chen*, Sylvia L. Herbert*, Haimin Hu, Ye Pu, Jaime F. Fisac, Somil Bansal, SooJean Han, Claire J. Tomlin
\thanks{This research is supported by ONR under the Embedded Humans MURI (N00014-16-1-2206). The research of S. Herbert has received funding from the NSF GRFP and the UC Berkeley Chancellor's Fellowship Program.}
\thanks{* Both authors contributed equally to this work.}
\thanks{M. Chen is with the School of Computing Science, Simon Fraser University. mochen@cs.sfu.ca}
\thanks{S. Herbert is with the Department of Mechanical and Aerospace Engineering, UC San Diego. sherbert@ucsd.edu}
\thanks{H. Hu and J. Fisac are with the Department of Electrical Engineering, Princeton University. \{haiminh, jfisac\}@princeton.edu}
\thanks{Y. Pu is with the Department of Electrical and Electronic Engineering, University of Melbourne. ye.pu@unimelb.edu.au}
\thanks{S. Bansal and C. Tomlin are with the Department of Electrical Engineering and Computer Sciences, UC Berkeley. \{somil, tomlin\}@berkeley.edu}
\thanks{S. Han is with the Control and Dynamical Systems program, California Institute of Technology. soojean@caltech.edu}
\thanks{A preliminary version of this paper was published in \cite{herbert2017fastrack}.}
}

\begin{document}
\maketitle
\thispagestyle{empty}
\pagestyle{empty}

\begin{abstract}
Real-time, guaranteed safe trajectory planning is vital for navigation in unknown environments. However, real-time navigation algorithms typically sacrifice robustness for computation speed. Alternatively, provably safe trajectory planning tends to be too computationally intensive for real-time replanning. We propose FaSTrack, Fast and Safe Tracking, a framework that achieves both real-time replanning and guaranteed safety. In this framework, real-time computation is achieved by allowing any trajectory planner to use a simplified \textit{planning model} of the system. The plan is tracked by the system, represented by a more realistic, higher-dimensional \textit{tracking model}. We precompute the tracking error bound (TEB) due to mismatch between the two models and due to external disturbances. We also obtain the corresponding tracking controller used to stay within the TEB. The precomputation does not require prior knowledge of the environment. We demonstrate FaSTrack using Hamilton-Jacobi reachability for precomputation and three different real-time trajectory planners with three different tracking-planning model pairs.
\end{abstract}

\section{Introduction}

\begin{figure}
 	\centering
 	\includegraphics[width=\columnwidth]{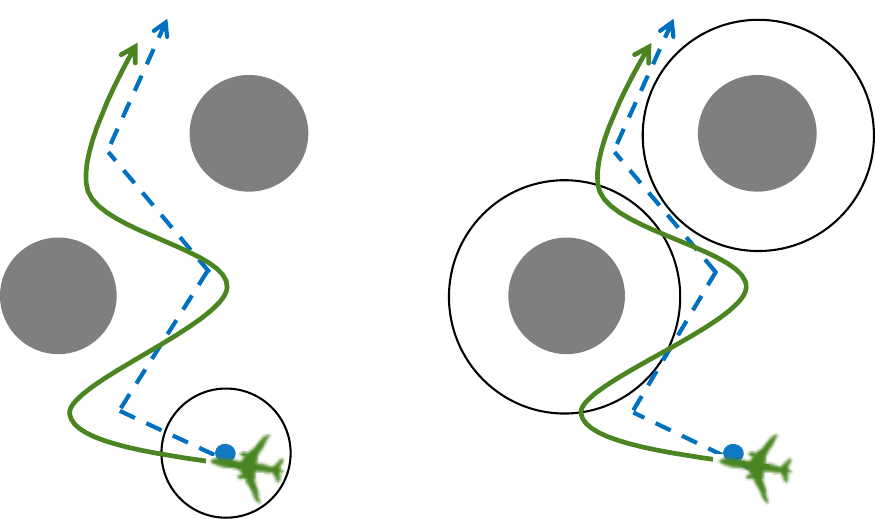}
 	\caption{Left: A planning algorithm uses a fast but simple model (blue disk), to plan around obstacles (gray disks). The more complicated tracking model (green plane) tracks the path. By using FaSTrack the autonomous system is guaranteed to stay within some TEB (black circle). Right: Safety can be guaranteed by planning with respect to obstacles augmented by the TEB (large black circles).}
 	\label{fig:chasing}
 \end{figure}

In autonomous dynamical systems, safety and real-time planning are both crucial for many applications.
This is particularly true when environments are \textit{a priori} unknown, because replanning based on updated information about the environment is often necessary.
However, achieving safe navigation in real time is difficult for many common dynamical systems due to the computational complexity of generating and formally verifying the safety of dynamically feasible trajectories.
To achieve real-time planning, many algorithms use highly simplified model dynamics or kinematics to create a nominal trajectory that is then tracked by the system using a feedback controller such as a linear quadratic regulator (LQR).  These nominal trajectories may not be dynamically feasible for the true autonomous system, resulting in a tracking error between the planned path and the executed trajectory.
 This concept is illustrated in Fig. \ref{fig:chasing}, where the path was planned using a simplified planning model, but the real dynamical system cannot track this path exactly.
Additionally, external disturbances (e.g. wind) can be difficult to account for using real-time planning algorithms, causing another source of tracking error.
These tracking errors can lead to dangerous situations in which the planned path is safe, but the actual system trajectory enters unsafe regions.  Therefore, real-time planning is achieved at the cost of guaranteeing safety.  Common practice techniques augment obstacles by an ad hoc safety margin, which may alleviate the problem but is performed heuristically and therefore does not guarantee safety.

To attain fast planning speed while maintaining safety, we propose the modular framework FaSTrack: Fast and Safe Tracking.
FaSTrack also allows planning algorithms to use a simplified model of the system in order to operate in real time using augmented obstacles.
However, in FaSTrack, the obstacle augmentation bound is rigorously computed and comes with a corresponding optimal tracking controller.
Together this \textit{tracking error bound} (TEB) and controller guarantee safety for the autonomous system as it tracks the simplified plans (see Fig. \ref{fig:chasing}, right).
We compute this bound and controller by modeling the navigation task as a pursuit-evasion game between a sophisticated \textit{tracking model} (pursuer) and the simplified \textit{planning model} of the system (evader).
The tracking model accounts for complex system dynamics as well as bounded external disturbances, while the simple planning model enables the use of real-time planning algorithms.
Offline, the pursuit-evasion game between the two models can be analyzed using any suitable method \cite{Mitchell05, SinghChenEtAl2018, royo2018classification} to produce a TEB.
To provide a TEB for all states of the planning model through solving a single pursuit-evasion game and to reduce the problem dimensionality, TEB computations are performed using the relative dynamics between the two models.

This results in a \textit{tracking error function} that maps the initial relative state between the two models to the TEB: the maximum possible relative distance that could occur over time.
The TEB can be thought of as a ``safety bubble" around the planning model of the system that the tracking model of the system is guaranteed to stay within.

The resulting TEB from this precomputation may converge to an invariant set (i.e. the planning model cannot move arbitrarily far away from the tracking model when both act optimally), or may result in a time-varying set.
Since the planning model can be designed by the user, typically one can select a model such that the computation converges.
However, there may be cases in which convergence does not occur (i.e. even when acting optimally the autonomous sytem cannot keep up with the planning model used by the path or trajectory planning algorithm).
In these cases we can instead compute a time-varying TEB.
Intuitively, a time-varying TEB means that as time progresses the tracking error bound increases by a known amount.

Because the tracking error is bounded in the relative state space, we can precompute and store the \textit{optimal tracking controller} that maps the real-time relative state to the optimal tracking control action for the tracking model to pursue the planning model.
The offline computations are \textit{independent} of the path planned in real time.

Online, the autonomous system senses local obstacles, which are then augmented by the TEB to ensure that no potentially unsafe paths can be computed.
Next, any chosen path or trajectory planning algorithm uses the simplified planning model and the local environment to determine the next planning state.
The autonomous system (represented by the tracking model) then finds the relative state between itself and the next desired state.
If this relative state is nearing the TEB then it is plugged into the optimal tracking controller to find the instantaneous optimal tracking control of the tracking model required to stay within the error bound; otherwise, any tracking controller may be used. In this sense, FaSTrack provides a \emph{least-restrictive} control law.
This process is repeated for as long as the planning algorithm (rapidly-exploring random trees, model predictive control, etc.) is active.

FaSTrack is modular, and can be used with any method for computing the TEB together with any existing path or trajectory planning algorithms.
Any feature of the planning algorithm, such as ability to account for time-varying obstacles, is inherited when used in the FaSTrack framework.
This enables motion planning that is real-time, guaranteed safe, and dynamically accurate.
FaSTrack was first introduced in \cite{herbert2017fastrack}, and is generalized here in a number of important ways.

First, we adopt definitions in \cite{SinghChenEtAl2018} to refine the notion of relative system to be much more general; in particular, the new definition of relative state allows the pairing of a large class of tracking and planning models.

Next, we introduce and prove the time-varying formulation of FaSTrack, which uses time-varying tracking controllers to provide a time-varying TEB (tvTEB).
This has significant practical impact, since the ability to obtain a tvTEB means that the FaSTrack framework does not depend on the convergence of pursuit-evasion games, which is in general not guaranteed and computationally expensive to check.

Furthermore, for both the time-invariant and time-varying cases, we provide mathematical proofs.
Lastly, we demonstrate the FaSTrack framework using three different real-time planning algorithms that have been ``robustified" by precomputing the TEB and tracking controller to demonstrate our framework.

Precomputation of the TEB and tracking control law for each planning-tracking model pair is done in this paper by solving a Hamilton-Jacobi (HJ) partial differential equation (PDE), which provides globally optimal solutions for general nonlinear systems.
We encourage readers to refer to \cite{SinghChenEtAl2018} for TEB computations using SOS optimization, which can sometimes be more scalable, but provides locally optimal solutions for systems with polynomial dynamics.
The planning algorithms used in our numerical examples are the fast sweeping method (FSM) \cite{Takei2013}, rapidly-exploring random trees (RRT) \cite{Kuffner2000,Kavraki1996}, and model predictive control (MPC) \cite{Qin2003,Zhang2017}.

In the three examples, we also consider different tracking and planning models, one of which utilizes the refined definition of relative system, and another involving a tvTEB.
In the simulations, the system travels through a static environment with constraints defined, for example, by obstacles, while experiencing disturbances.
The constraints are only fully known through online sensing (e.g. once obstacles are within the limited sensing region of the autonomous system).
By combining the TEB with real-time planning algorithms, the system is able to safely plan and track a trajectory through the environment in real time.
\section{Related Work \label{sec:relatedwork}}
Motion planning is an active research area in the controls and robotics communities \cite{Hoy2015}.  In this section we will discuss past work on path, kinematic, and dynamic planning.
A major current challenge is to find an intersection of robust and real-time planning for general nonlinear systems.
Sample-based planning methods like rapidly-exploring random trees (RRT) \cite{Kuffner2000}, probabilistic road maps (PRM) \cite{Kavraki1996}, fast marching tree (FMT) \cite{Janson2015}, fast sweeping method \cite{Takei2013} and many others \cite{Richter2016, Karaman2011, Kobilarov2012} can find collision-free paths through known or partially known environments. While extremely effective in a number of use cases, these algorithms are not designed to be robust to model uncertainty or disturbances, and may not even use a dynamic model of the system in the first place.
Motion planning for kinematic systems can also be accomplished through online trajectory optimization using methods such as TrajOpt \cite{Schulman2013} and CHOMP \cite{Ratliff2009}. These methods can work extremely well in many applications, but are generally challenging to implement in real time for nonlinear dynamic systems.

Model predictive control (MPC) has been a very successful method for dynamic trajectory optimization \cite{Qin2003}.  However, combining speed, safety, and complex dynamics is a difficult balance to achieve. Using MPC for robotic and aircraft systems typically requires model simplification to take advantage of linear programming or mixed integer linear programming \cite{Vitus2008, Zeilinger2011, Richter2012}; robustness can also be achieved in linear systems \cite{Richards2006, DiCairano2016}. Nonlinear MPC is most often used on systems that evolve more slowly over time \cite{Diehl2002, Schildbach2016}, with active work to speed up computation \cite{Diehl2009, Neunert2016}. Adding robustness to nonlinear MPC is being explored through algorithms based on min-max formulations and tube MPCs that bound output trajectories around a nominal path (see \cite{Hoy2015} for references).

There are other methods of dynamic trajectory planning that manage to cleverly skirt the issue of solving for optimal trajectories online.  One such class of methods involve motion primitives \cite{Gillula2010, Dey2016}. Other methods include making use of safety funnels \cite{Majumdar2017}, or generating and choosing random trajectories at waypoints \cite{Kalakrishnan2011, Schwesinger2013}. The latter methods have been implemented successfully in many scenarios, but can be risky in their reliance on finding combinations of pre-computed or randomly-generated safe trajectories.

One notable real-time planning method that also involves robustness guarantees is given by \cite{KousikVaskovEtAl2017}, in which a forward reachable set for a high-fidelity model of the system is computed offline and then used to prune motion plans generated online using a low-fidelity model.
The approach relies on an {\em assumed} model mismatch bound; therefore our work has potential to complement works such as \cite{KousikVaskovEtAl2017} by providing the TEB as well as a corresponding feedback tracking controller.

Recent work has considered using offline Hamilton-Jacobi analysis to guarantee tracking error bounds, which can then be used for robust trajectory planning \cite{Bansal2017}.

A class of closely-related techniques define safe tubes around a nominal dynamic trajectories by constructing control-Lyapunov functions, which tend to be very difficult to compute \cite{Burridge1999}.
In recent years, methods involving using contraction theory and numerous optimization techniques have enabled computation of conservative approximations of control-Lyapunov functions in the context of robust trajectory tracking \cite{Parrilo2000,Majumdar2017,Singh2017,Ahmadi2017}.

Finally, some online control techniques can be applied to trajectory tracking with constraint satisfaction. For control-affine systems in which a control barrier function can be identified, it is possible to guarantee forward invariance of the desired set through a state-dependent affine constraint on the control, which can be incorporated into an online optimization problem, and solved in real time \cite{Ames2014}. This can be seen as a method for smoothly blending some of the above safety results with other, performance-based objectives.
There are also motion planning methods designed to be robust. Control barrier functions \cite{Xu2015, Ames2014} place inequality constraints in the control input that allow for dynamic trajectory planning as a quadratic program. New work in planning using contraction theory works by forming safe tubes online around a nominal dynamic trajectory \cite{Singh2017}. Offline planners like HJ analysis can find control policies and guarantees for nonlinear systems that avoid obstacles and are robust to bounded disturbances \cite{Mitchell05}. However, this method can only approach real-time speed for very low-dimensional (1D-2D) systems, despite recent work in alleviating the dimensionality hurdle \cite{Chen2016a, Chen2016b}.

The work presented in this paper differs from the robust planning methods above because FaSTrack is designed to be modular and easy to use in conjunction with any path or trajectory planner. Additionally, FaSTrack can handle bounded external disturbances (e.g. wind) and work with both known and unknown environments with obstacles.
\section{Problem Formulation \label{sec:formulation}}
FaSTrack is a modular framework to plan and track a trajectory (or path converted to a trajectory) online and in real time.
Planning is done using a relatively simple model of the system, called the \textit{planning model.
The planning model and algorithm should be chosen to allow real-time planning.}

On the other hand, tracking is achieved by a \textit{tracking model} that more accurately represents the autonomous system.
Practically, the tracking model should be chosen to consider factors such as higher-order dynamics and disturbances.
Under the FaSTrack framework, a tracking error bound (TEB) is computed to account for the mismatch between the planning and tracking models to realize the benefits of both using a simplified model and a higher-fidelity model.
Although this paper focuses on using the Hamilton-Jacobi method to compute time-invariant and time-varying TEBs, in general any method can be used under the FaSTrack framework.
For example, the authors in \cite{SinghChenEtAl2018} use SOS optimization to achieve better computational scalability.

The environment may contain static obstacles that are \textit{a priori} unknown and can be observed by the system within a limited sensing range.
In this section we define the tracking and planning models, as well as the goals of the paper.

\subsection{Tracking  Model}
The tracking model is a relatively accurate and typically higher-dimensional representation of the autonomous system dynamics.
Let $\tstate\in \tset\subseteq\R^{n_s}$ represent the states of the tracking model.
The evolution of the tracking model dynamics satisfies the following ordinary differential equation (ODE):

\begin{equation}
\begin{aligned}
\label{eq:tdyn}
\frac{d\tstate}{d\tvar} = \dot{\tstate} = \tdyn(\tstate(\tvar), \tctrl(\tvar), \dstb(\tvar)), \tvar \in [0, \thor], \\
\tstate(\tvar) \in \tset, \tctrl(\tvar) \in \tcset, \dstb(\tvar) \in \dset,
\end{aligned}
\end{equation}

We assume that the tracking model dynamics $\tdyn : \tset\ \times\ \tcset \times \dset \rightarrow \tset$ is Lipschitz continuous in the system state $\tstate$ for a fixed control and disturbance functions $\tctrl(\cdot), \dstb(\cdot)$.
At every time $t$, the control $\tctrl$ is constrained by the compact set $\tcset\subseteq \R^{n_{u_s}}$, and the disturbance $\dstb$ by the compact set $\dset\subseteq \R^{n_d}$.
Furthermore, the control function $\tctrl(\cdot)$ and disturbance function $\dstb(\cdot)$ are measurable functions of time:

\begin{align}
\tctrl(\cdot) \in \tcfset &:= \{\phi: [0, \thor] \rightarrow \tcset, \phi(\cdot) \text{ is measurable}\},\\
\dstb(\cdot) \in \dfset &:= \{\phi: [0, \thor] \rightarrow \dset, \phi(\cdot) \text{ is measurable}\}.
\end{align}

\noindent where $\tcfset$ and $\dfset$ represent the set of functions that respectively satisfy control and disturbance constraints at all times.
Under these assumptions there exists a unique trajectory solving (\ref{eq:tdyn}) for a given $\tctrl(\cdot) \in \tcfset, \dstb(\cdot)\in\dfset$ \cite{Coddington84}. The trajectories of (\ref{eq:tdyn}) that solve this ODE will be denoted as $\ttraj(\tvar; \tstate, \tvar_0, \tctrl(\cdot), \dstb(\cdot))$, where $\tvar_0,\tvar \in [0, \thor]$ and $\tvar_0 \leq \tvar$. This trajectory notation represents the state of the system at time $\tvar$, given that the trajectory is initiated at state $\tstate$ and time $\tvar_0$ and the applied control and disturbance functions are $\tctrl(\cdot)$ and disturbance $\dstb(\cdot)$ respectively.
These trajectories will satisfy the initial condition and the ODE (\ref{eq:tdyn}) almost everywhere:

\begin{align*}
&\frac{d}{d\tvar}\ttraj(\tvar; \tstate_0, \tvar_0, \tctrl(\cdot), \dstb(\cdot)) = \\ &\qquad \tdyn(\ttraj(\tvar; \tstate_0, \tvar_0, \tctrl(\cdot), \dstb(\cdot)), \tctrl(\tvar), \dstb(\cdot)), \\
&\ttraj(\tvar_0; \tstate_0, \tvar_0, \tctrl(\cdot), \dstb(\cdot)) = \tstate_0.
\end{align*}

Let $\tgoal \subset \tset$ represent the set of goal states, and $\tconstr \subset \tset$ represent state constraints for all time.
Often, $\tconstr$ represents the complement of obstacles that the system must avoid.

\example{We introduce a running example for illustration throughout the paper. In this example a car will have to navigate through an environment with a priori unknown obstacles ($\tconstr^\complement$) towards a goal ($\tgoal$). The tracking model of the car is represented by the following five-dimensional dynamics:
\begin{equation}
\label{eq:5Ddyn}
\begin{bmatrix}
\dot x\\
\dot y\\
\dot\theta\\
\dot v\\
\dot \omega
\end{bmatrix} =
\begin{bmatrix}
v \cos \theta + \dstb_x\\
v \sin \theta + \dstb_y\\
\omega \\
a + \dstb_a\\
\alpha + \dstb_\alpha
\end{bmatrix},
\end{equation}
\noindent where $(x,y,\theta)$ represent the pose (position and heading) of the 5D car model, and $(v, \omega)$ are the speed and turn rate.
The control of the 5D model consists of the linear and angular acceleration, $(a, \alpha)$, and the disturbances are $(\dstb_x, \dstb_y, \dstb_a, \dstb_\alpha)$.  The model parameters are chosen to be $a \in [-0.5, 0.5]$, $|\alpha|\le 6$, $|\dstb_x|, |\dstb_y|, |\dstb_\alpha|\le0.02$, $|\dstb_a| \le 0.2$.
}

\subsection{Planning Model \label{sec:planning_model}}
The planning model is a simpler, lower-dimensional model of the system.
Replanning is necessary for navigation in unknown environments, so the planning model is typically constructed by the user so that the desired planning algorithm can operate in real time.

Let $\pstate$ represent the state of the planning model, and let $\pctrl$ be the control.
We assume that the planning state $\pstate \in \pset$ are a subset of the tracking state $\tstate \in \tset$, so that $\pset$ is a subspace within $\tset$.
This assumption is reasonable since a lower-fidelity model of a system typically involves a subset of the system's states.
The dynamics of the planning model satisfy

\begin{align}
\label{eq:pdyn}
\frac{d\pstate}{d\tvar} = \dot{\pstate} = \pdyn(\pstate, \pctrl), \tvar \in [0, \thor], \quad \pstate \in \pset, \pctrl \in \pcset,
\end{align}

\noindent with the analogous assumptions on continuity and compactness as those for \eqref{eq:tdyn}.

Note that the planning model does not include a disturbance input.
This is a key feature of FaSTrack: the treatment of disturbances is only necessary in the tracking model, which is modular with respect to any planning method. Therefore we can and will assume that the planning model (and the planning algorithm) do not consider disturbances. This allows the algorithm to operate efficiently without the need to consider robustness. If the planning algorithm does consider disturbances then the added robustness of FaSTrack may result in added conservativeness.

Let $\goal \subset \pset$ and $\constr \subset \pset$ denote the projection of $\tgoal$ and $\tconstr$ respectively onto the subspace $\pset$.
We will assume that $\constr$ is \textit{a priori} unknown, and must be sensed as the autonomous system moves around in the environment.
Therefore, for convenience, we denote the currently known, or ``sensed'' constraints as $\constrSense(t)$.
Note that $\constrSense(t)$ depends on time, since the system may gather more information about constraints in the environment over time.
In addition, as described throughout the paper, we will augment $\constrSense(t)$ according to the TEB between the tracking and planning models.
We denote the augmented obstacles as $\constrAug(t)$.

\example{For efficient planning use a simpler 3D model with the following dynamics:
\begin{equation}
\dot \pstate =
\begin{bmatrix}
\dot {\hat x}\\
\dot {\hat y}\\
\dot {\hat \theta}
\end{bmatrix}
=
\begin{bmatrix}
\hat v \cos \hat\theta\\
\hat v \sin \hat\theta\\
\hat \omega
\end{bmatrix},
\end{equation}
\noindent where $(\hat x, \hat y, \hat\theta)$ represent the pose (position and heading) of the 3D car model. Here the speed $\hat v$ is a constant, and the turn rate $\hat \omega$ is the control. The planning model must reach its goal $\goal$ while avoiding obstacles represented by $\constrAug(t)$.  The model parameters are chosen to be $\hat v = 0.1$, $|\hat\omega|\le 1.5$.
}

\subsection{Goals and Approach}
Given system dynamics in \eqref{eq:tdyn}, initial state $\tstate_0$, goal states $\tgoal$, and constraints $\tconstr$ such that $\constr$ is \textit{a priori} unknown and determined in real time, we would like to steer the system to $\tgoal$ with formally guaranteed satisfaction of $\tconstr$, despite any disturbances the system may experience.

To achieve this goal, FaSTrack decouples the formal safety guarantee from the planning algorithm.
Instead of having the system, represented by the tracking model, directly plan trajectories towards $\tgoal$, the system (represented by the tracking model) ``chases'' the planning model of the system, which may use any planning algorithm to obtain trajectories in real time.
The autonomous system is guaranteed to stay within the TEB relative to the planning model, as we will prove in Prop. \ref{prop:nonconv}, and arrive at its goal as long as the TEB is entirely within the goal.
Therefore, we set $\contrgoal$ to be the projection of $\tgoal$ onto the subspace $\pset$ and contracted by one TEB.
When the planning algorithm reaches $\contrgoal$, we know that the autonomous system will be within $\tgoal$.
Safety is formally guaranteed through precomputation of the TEB along with a corresponding optimal tracking controller, in combination with augmentation of constraints based on this TEB.
An illustration of our framework is shown in Fig. \ref{fig:chasing}.
\section{General Framework \label{sec:framework}}
\begin{figure}[h!]
	\centering
	\includegraphics[width=\columnwidth]{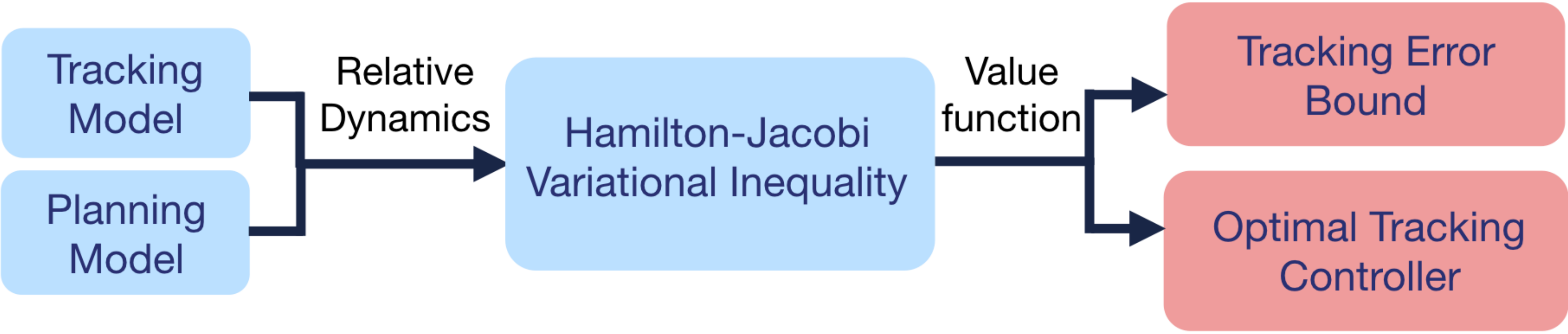}
	\caption{Offline framework. Output of offline framework shown in red.}
	\label{fig:fw_offline}
\end{figure}

\begin{figure}[h!]
	\centering
	\includegraphics[width=\columnwidth]{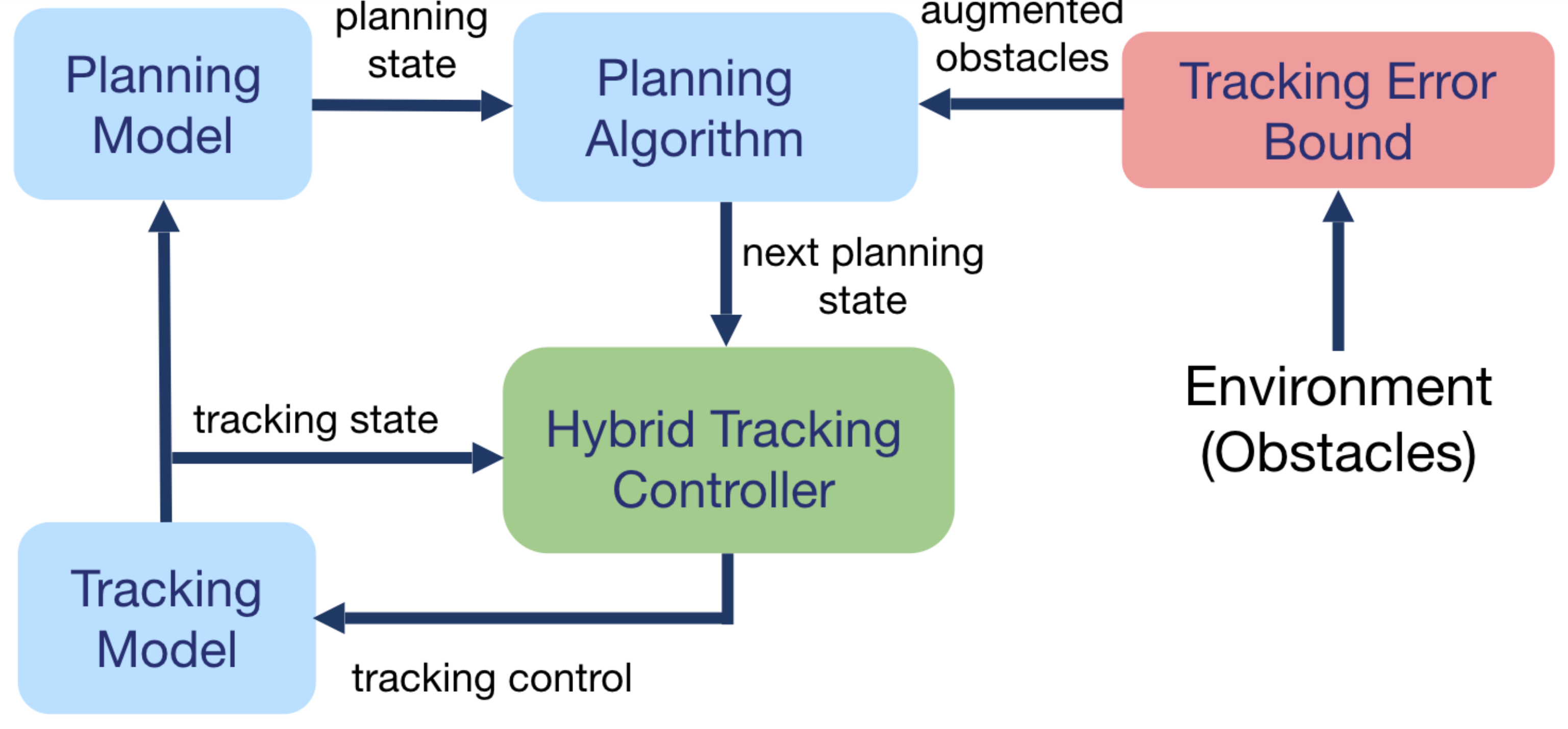}
	\caption{Online framework. Components from offline computation shown in red.}
	\label{fig:fw_online}
\end{figure}

\begin{figure}[h!]
	\centering
	\includegraphics[width=\columnwidth]{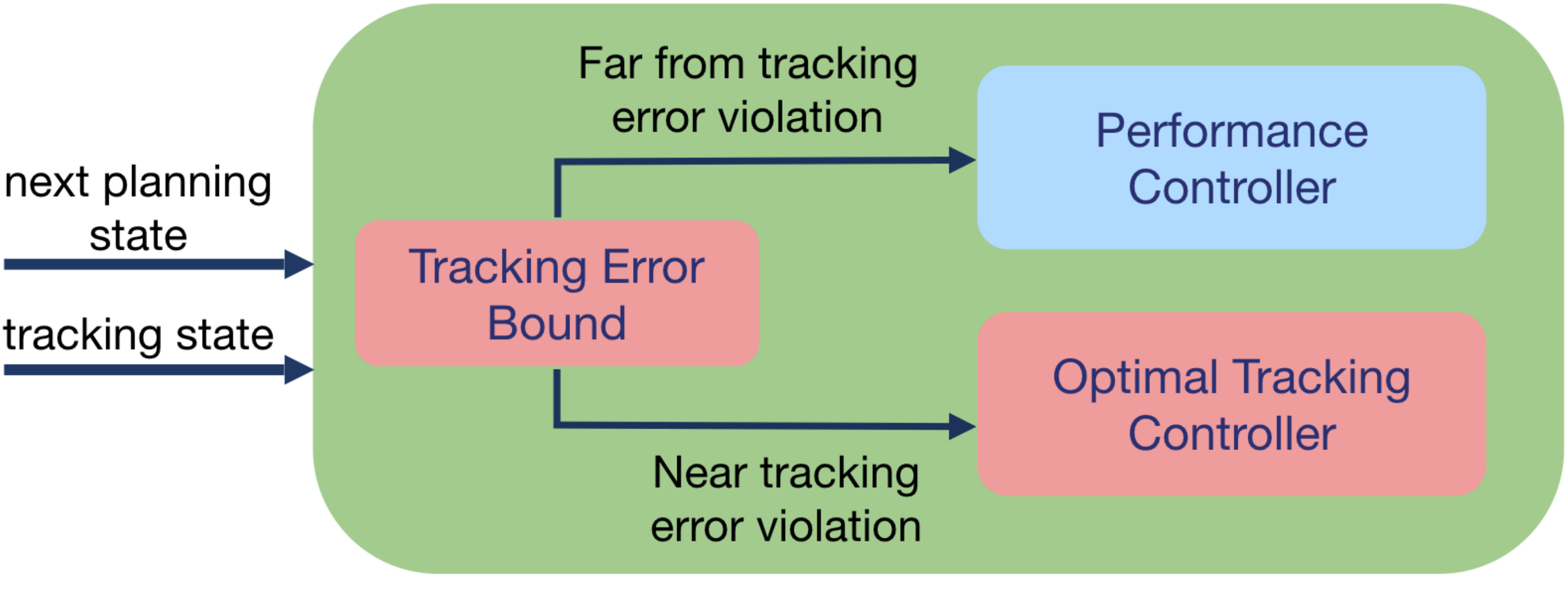}
	\caption{Hybrid controller. Components from offline computation are shown in red.}
	\label{fig:hybrid_ctrl}
\end{figure}

Details of the framework are summarized in Figs. \ref{fig:fw_offline}, \ref{fig:fw_online}, and \ref{fig:hybrid_ctrl}.
The purpose of the offline framework (Fig. \ref{fig:fw_offline}) is to generate a TEB and corresponding optimal tracking controller that can be quickly and easily used by the online framework.
The planning and tracking model dynamics are used in the reachability precomputation (described in sec. \ref{sec:precomp}), whose solution is a value function that acts as the TEB function/look-up table.
The gradients of the value function comprise the optimal tracking controller function/look-up table.
These functions are independent of the online computations and environment -- they depend only on the \textit{relative system state} and dynamics between the planning and tracking models, not on absolute states along the trajectory at execution time.

Online, we start in the bottom-left corner of Fig. \ref{fig:fw_online} to determine the tracking model's initial state (i.e. autonomous system's initial state). Based on this we initialize the planning model such that the tracking model is within the TEB relative to the planning model. The state of the planning model is entered into a planning algorithm.
Another input to the planning algorithm is the set of augmented constraints $\constrAug$.  These are acquired by updating constraints $\constr$ and accordingly updating the sensed constraints $\constrSense$ in the environment.  This can be done, for example, by sensing the environment for obstacles. Next, $\constrSense$ is augmented  by the precomputed TEB using the Minkowski difference to produce the augmented constraints $\constrAug$. \footnote{For a faster computation we typically expand each obstacle by the maximum distance of the TEB in each dimension as a conservative approximation.}

In terms of obstacles in the environment, augmenting the constraints by this margin can be thought of as equivalent to wrapping the planning model of the system with a ``safety bubble''.
The planning algorithm takes in the planning model state and augmented constraints, and then outputs a next desired state for the planning model towards $\contrgoal$.
The hybrid tracking controller block takes in this next planning model state along with the current state of the tracking model.
Based on the relative state between these two models, the hybrid tracking controller outputs a control signal to the autonomous system.
The goal of this control is to make the autonomous system track the desired planning state as closely as possible.
This cycle continues as the planning algorithm moves towards $\contrgoal$.

The hybrid tracking controller is expanded in Fig. \ref{fig:hybrid_ctrl} and consists of two controllers: an \textit{optimal tracking controller }(also referred to as the safety controller) and a \textit{performance controller}.
In general, there may be multiple safety and performance controllers depending on various factors such as observed size of disturbances, but for simplicity we will just consider one safety and one performance controller in this paper.
The optimal tracking controller consists of a function (or look-up table) computed offline by solving a HJ variational inequality (VI) \cite{Fisac15}, and guarantees that the TEB is not violated, despite the worst-case disturbance and worst-case planning control.
Although the planning model in general does not apply the worst-case planning control, assuming the worst allows us to obtain a \textit{trajectory-independent} TEB and an optimal tracking controller that is guaranteed safe.
Note that the computation of the value function and optimal tracking controller is done offline; during online execution, the table look-up operation is computationally inexpensive.
The memory required to store the value function is the same as the memory required to compute it. This amount depends on the grid resolution, whether the value function is finite time horizon or infinite time horizon, and in the former case, the length of the time horizon. We have specified the RAM usage in each of our examples in Section \ref{sec:results}.

When the system is close to violating the TEB, the optimal tracking controller must be used to prevent the violation.
On the other hand, when the system is far from violating the TEB, any controller (such as one that minimizes fuel usage), can be used.
This control is used to update the autonomous system's state, and the process repeats.
In the following sections we will first explain the precomputation steps taken in the offline framework.
We will then walk through the online framework.
Finally, we will present three numerical examples.
\section{Offline Computation \label{sec:precomp}}
The offline computation begins with setting up a pursuit-evasion game \cite{Tomlin00,Mitchell05} between the tracking model and the planning model of the system.
In this game, the tracking model will try to ``capture" the planning model, while the planning model is doing everything it can to avoid capture.
In reality the planning algorithm is typically not actively trying to avoid the tracking model, but this allows us to account for worst-case scenarios and more crucially, ensure that the TEB is \textit{trajectory-independent}.
If both systems are acting optimally in this way, we can determine the
maximum possible tracking error between the two models, which is captured by the value function obtained from solving a Hamilton-Jacobi variational inequality, as described below.

\subsection{Relative System Dynamics\label{subsec:rel_sys}}
To determine the relative distance over time, we must first define the relative system derived from the tracking (\ref{eq:tdyn}) and planning (\ref{eq:pdyn}) models.
The relative system is obtained by fixing the planning model to the origin and finding the dynamics of the tracking model relative to the planning model.
Defining $\rstate$ to be the relative system state, we write

\begin{equation}
\label{eq:rstate}
\rstate = \rtrans(\tstate,\pstate)(\tstate - \ptmat\pstate)
\end{equation}

\noindent where $\ptmat$ matches the common states of $\tstate$ and $\pstate$ by augmenting the state space of the planning model.
The relative system states $\rstate$ represent the tracking system states relative to the planning states.
The function $\rtrans$ is a linear transform that simplifies the relative system dynamics to be of the form

\begin{equation}
\label{eq:rdyn}
\dot\rstate = \rdyn(\rstate, \tctrl, \pctrl, \dstb),
\end{equation}

\noindent which only depends on the relative system state $r$.
A transform $\rtrans$ that achieves the relative system dynamics in the form of \eqref{eq:rdyn} is often the identity map or the rotation map when the autonomous system is a mobile robot; therefore, in this paper, we assume that a suitable $\rtrans$ is available.
For general dynamical systems, it may be difficult to determine $\rtrans$; a catalog of tracking and planning models with suitable transforms $\rtrans$, as well as a more detailed discussion of $\rtrans$, can be found in \cite{SinghChenEtAl2018}.

In addition, we define the error state $\estate$ to be the relative system state \textit{excluding} the absolute states of the tracking model, and the auxiliary states $\astate$ to be the relative system state \textit{excluding} the error state.
Hence, $\rstate = \begin{bmatrix} \estate, \astate \end{bmatrix}^\intercal$.

\example{We must determine the relative system state between our 5D tracking and 3D planning models of the car. We define the relative system state to be $(x_r, y_r, \theta_r, v, \omega)$, such that the error state $\estate=[x_r, y_r, \theta_r]^\intercal$ is the position and heading of the 5D model in the reference frame of the 3D model, and the auxiliary state $\astate = [v, \omega]^\intercal$ represents the speed and turn rate of the 5D model.
	The relative system state $\rstate = \begin{bmatrix} \estate, \astate \end{bmatrix}^\intercal$, tracking model state $\tstate$, and planning model state $\pstate$ are related through $\rtrans$ and $\ptmat$ as follows:
	\begin{equation}
	\label{eq:5D_and_3D_err_state}
	\underbrace{
		\begin{bmatrix}
		x_r\\
		y_r\\
		\theta_r \\
		v \\
		\omega
		\end{bmatrix}
	}_\rstate
	=
	\underbrace{
		\begin{bmatrix}
		\begin{bmatrix}
		\cos\hat\theta & \sin\hat\theta \\
		-\sin\hat\theta & \cos\hat\theta
		\end{bmatrix} & \mathbf{0_{2\times 3}} \\
		\mathbf{0_{3\times 2}} & \mathbf I_3
		\end{bmatrix}
	}_\rtrans
	\Bigg(
	\underbrace{
		\begin{bmatrix}
		x\\
		y\\
		\theta \\
		v \\
		\omega
		\end{bmatrix}
	}_\tstate -
	\underbrace{
		\begin{bmatrix}
		\mathbf I_3 \\
		\mathbf{0_{2\times 3}}
		\end{bmatrix}
	}_\ptmat
	\underbrace{
		\begin{bmatrix}
		\hat x\\
		\hat y \\
		\hat \theta \\
		\end{bmatrix}
	}_\pstate
	\Bigg),
	\end{equation}
	\noindent where $\mathbf 0, \mathbf I$ denote the zero and identity matrices of the indicated sizes.
	Taking the time derivative, we obtain the following relative system dynamics:
	\begin{equation}
	\label{eq:5D_and_3D_rdyn}
	\dot \rstate =
	\begin{bmatrix}
	\dot \estate\\
	\dot \astate
	\end{bmatrix}
	=
	\begin{bmatrix}
	\dot x_r\\
	\dot y_r\\
	\dot\theta_r\\
	\dot v\\
	\dot \omega
	\end{bmatrix}
	=
	\begin{bmatrix}
	- \hat v + v \cos \theta_r + \hat \omega y_r + \dstb_x\\
	v \sin \theta_r - \hat \omega x_r + \dstb_y\\
	\omega - \hat \omega \\
	a + \dstb_a\\
	\alpha + \dstb_\alpha
	\end{bmatrix}.
	\end{equation}
}

More examples of relative systems are in Section \ref{sec:results}.

\subsection{Formalizing the Pursuit-Evasion Game}
\begin{figure}
	\includegraphics[width=\columnwidth]{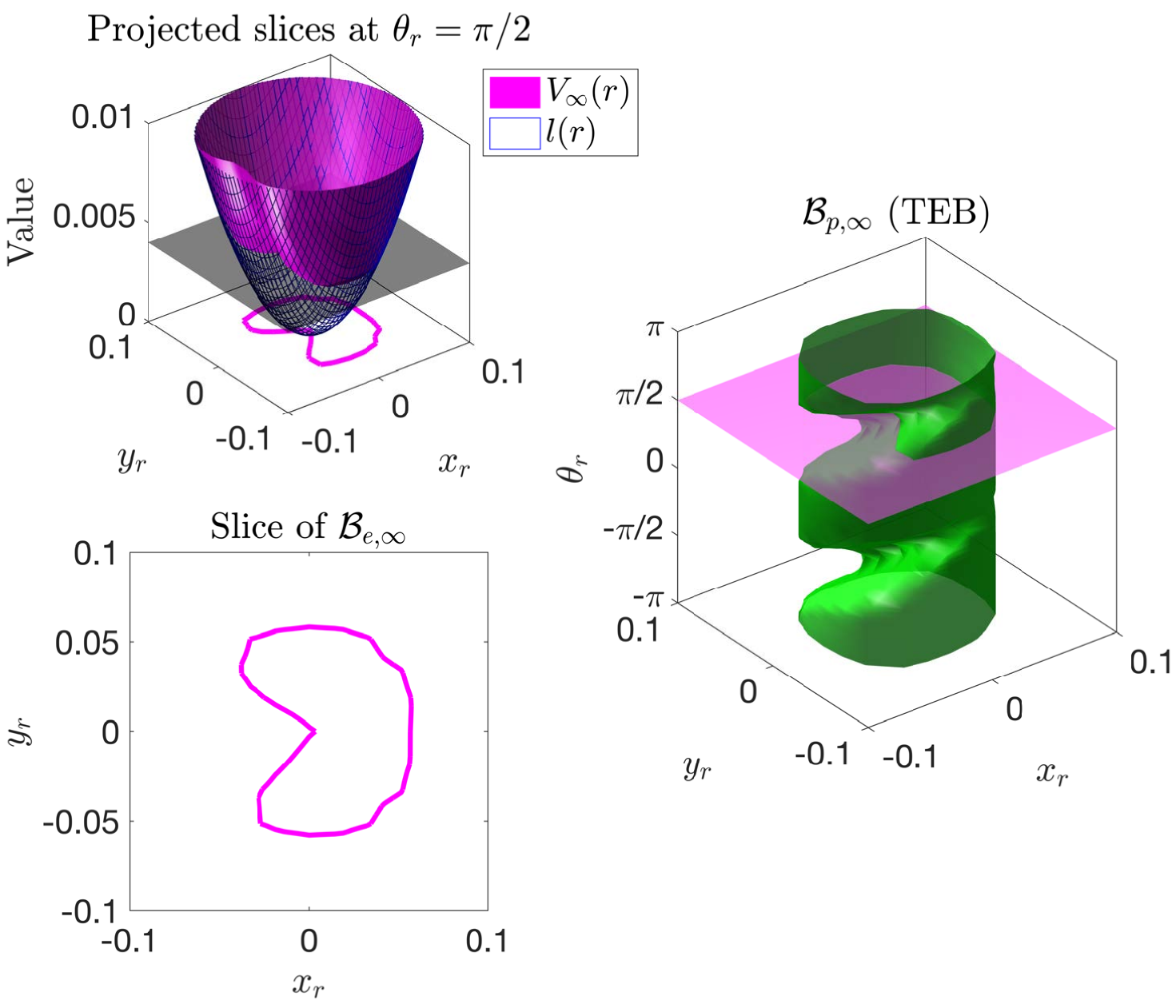}
	\caption{Value function and TEB for the running example in \eqref{eq:5D_and_3D_err_state} and \eqref{eq:5D_and_3D_rdyn}. Top Left: projected slice ($\theta_r = \pi/2$) of the error function (blue hatch) and converged value function (magenta). The minimum value $\underline\valfunc$ of the converged value function is marked by the black plane; the slice of the value function at $\underline\valfunc$ determines the TEB (pink set), also shown on the bottom left.
	Right: the full TEB (no longer projected) in the error states. Note that the slice shown on the bottom left corresponds to the slice marked by the magenta plane at $\theta_r = \pi/2$.}
	\label{fig:valfunc_illustration}
\end{figure}

Given the relative system dynamics between the tracking and planning models, we would like to compute a guaranteed TEB between these models.
This is done by first defining an error function $\errfunc(\rstate)$ in the relative state space.
One simple error function is the squared distance to the origin, which is shown in Fig. \ref{fig:valfunc_illustration} (top left, blue hatch surface), and is used when one is concerned only with the tracking error in position.

\begin{figure}
	\centering
ttt	\includegraphics[width=\columnwidth]{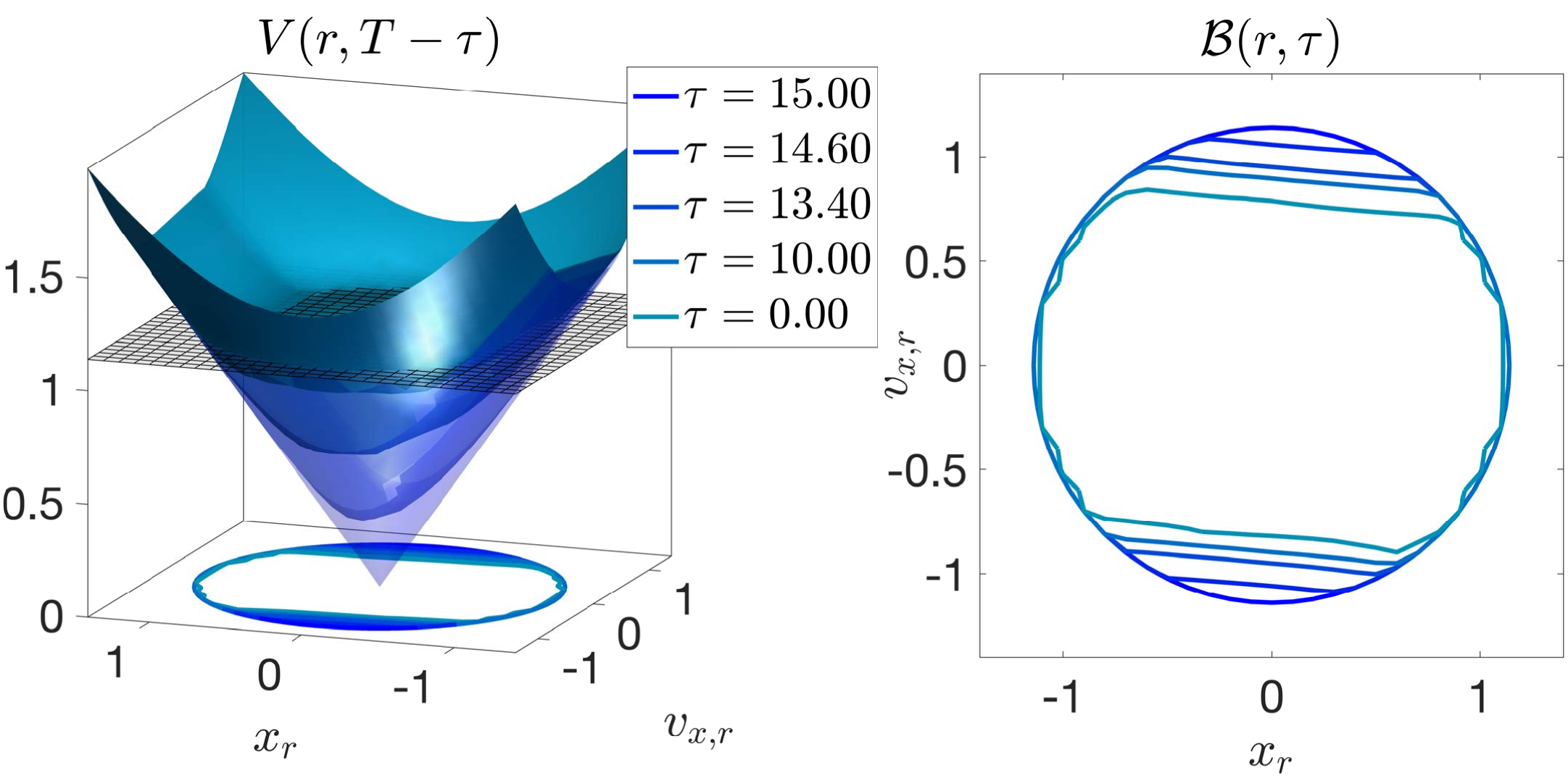}
	\caption{Time-varying value function (left) and TEBs (right) for the 8D quadrotor tracking 4D double integrator example in Section \ref{sec:resultsMPC}. The value function and the TEB varies with $\tau$, which represents time into the future.
  The size of the TEB increases with $\tau$ because the disturbance and planning control may drive the error states farther and farther from the origin over time.
  The error states shown are the relative position $x_r$ and velocity $v_{x,r}$.  Note that $\valfunc(\rstate,0) = \errfunc(\rstate)$.}
	\label{fig:vf_TEB:8D4D}
\end{figure}

When we would like to quantify the tracking error for more planning states (for example, error in angular orientation between the two models), the error function can be defined over these states as well.
For example, the error function seen in Fig. \ref{fig:vf_TEB:8D4D} is defined in both position and velocity space; Fig. \ref{fig:valfuncRRT} shows yet another error function defined using the one-norm of the displacement between the two models.
In our pursuit-evasion game, the tracking model tries to minimize the error, while the planning model and any disturbances experienced by the tracking model try to maximize.

Before constructing the pursuit-evasion game we must first define the method each player must use for making decisions.
We define a strategy for planning model as the mapping $\gamma_{\pstate} : \tcset \rightarrow \pcset$ that determines a planning control based on the tracking control. We restrict $\gamma$ to non-anticipative strategies $\gamma_{\pstate} \in \Gamma_\pstate(t)$, as defined in \cite{Mitchell05}.
We similarly define the disturbance strategy $\gamma_{\dstb}: \tcset \rightarrow \dset$, $\gamma_{\dstb} \in \Gamma_\dstb(t)$.

We compute the highest cost that this game will ever attain when both players are acting optimally.
This is expressed through the following value function:

\begin{align}
&V(\rstate,\thor)= \sup_{\gamma_{\pstate} \in \Gamma_\pstate(t), \gamma_{\dstb} \in \Gamma_\dstb(t)} \inf_{\tctrl(\cdot) \in \tcfset(t)} \big\{ \nonumber \\
&\qquad \max_{\tvar\in [0, \thor]} \errfunc\Big(\rtraj(\tvar; \rstate, 0, \tctrl(\cdot), \gamma_\pstate[\tctrl](\cdot), \gamma_\dstb[\tctrl](\cdot))\Big)\big\} \label{eq:valfunc}
\end{align}

The value function can be computed via existing methods in HJ reachability analysis \cite{Mitchell05, Fisac15}.
Adapting the formulation in \cite{Fisac15} and taking a convention of negative time in the backward reachability literature \cite{Chen2016DecouplingJournal, Chen2018}, we compute the value function by solving the HJ variational inequality

\begin{align}
\max \Big\{&\frac{\partial \tilde\valfunc}{\partial \tvar} + \min_{\tctrl\in\tset} \max_{\pctrl\in\pset, \dstb\in\dset} \nabla \tilde\valfunc \cdot \rdyn(\rstate, \tctrl, \pctrl, \dstb), \nonumber \\
&\qquad\errfunc(\rstate) - \tilde\valfunc(\rstate, \tvar)\Big\} = 0, \quad \tvar \in [-T, 0], \label{eq:HJVI} \\
&\tilde\valfunc(\rstate, 0) = \errfunc(\rstate), \nonumber
\end{align}

\noindent from which we obtain the value function, $\valfunc(\rstate, t) = \tilde\valfunc(\rstate, -t)$. There are many methods for solving this HJ variational inequality, including the level set toolbox \cite{Mitchell07c}.

If the planning model is ``close" to the tracking model and/or if the control authority of the tracking model is powerful enough to always eventually remain within some distance from the planning model, this value function will converge to an invariant solution for all time, i.e. $\valfunc_\infty(\rstate) := \lim_{\thor\rightarrow\infty} \valfunc(\rstate, \thor)$.
Because the planning model is user-defined, convergence can often be achieved by tuning the planning model.

However, there may be tracking-planning model pairs for which the value function does not converge.
In these cases the value function provides a finite time horizon, time-varying TEB, an example of which is shown in Fig. \ref{fig:vf_TEB:8D4D}.
Thus, even when convergence does not occur we can still provide time-varying safety guarantees.
In the rest of this paper, we will focus on the more general time-varying TEB case for clarity, and leave discussion of the time-invariant TEB case in the Appendix.
Our numerical examples will demonstrate both cases.

\example{We will set the cost to $\errfunc(\rstate) = x_r^2 + y_r ^2$, squared distance to the origin in position space. This means we would like the system to stay within an $(x_r,y_r)$ bound relative to the planning model, and ignore relative angle.  We initialize equation (\ref{eq:HJVI}) with this cost function and the relative system dynamics \eqref{eq:5D_and_3D_rdyn}.  We propagate the HJ variational inequality using the level set method toolbox until convergence or until we reach the planning horizon. In this case the value function converges to $\valfunc_\infty$ as seen in Fig. \ref{fig:vf_TEB:5D3D}.
}

In Section \ref{sec:proofs}, we formally prove that sublevel sets of $\valfunc(\rstate,\tvar)$ provide the corresponding time-varying TEBs $\TEB(t)$ for the finite time horizon case.
The analogous result for the time-invariant, infinite time horizon case is proven in the appendix.

The optimal tracking controller is obtained from the value function's spatial gradient \cite{Mitchell05, Fisac15, Chen2018}, $\deriv(\rstate, \tvar)$, as

\begin{align} \label{eq:opt_ctrl_fin}
\tctrl^*(\rstate, \tvar) = \arg\min_{\tctrl\in\tcset} \max_{\pctrl\in\pcset, \dstb\in\dset} \nabla\valfunc(\rstate, \tvar) \cdot \rdyn(\rstate,\tctrl,\pctrl,\dstb)
\end{align}

To ensure the relative system remains within the TEB, we also note that the optimal (worst-case) planning control $\pctrl^*$ and disturbance $\dstb^*$ can also obtained from $\deriv(\rstate, \tvar)$ as follows:

\begin{align} \label{eq:opt_dstb_fin}
\begin{bmatrix}
  \pctrl^* \\
  \dstb^*
\end{bmatrix} (\rstate, \tvar) = \arg \max_{\pctrl\in\pcset, \dstb\in\dset} \nabla\valfunc(\rstate, \tvar) \cdot \rdyn(\rstate,\tctrl^*,\pctrl,\dstb)
\end{align}

For system dynamics affine in the tracking control, planning control, and disturbance, the optimizations in \eqref{eq:opt_ctrl_fin} and \eqref{eq:opt_dstb_fin} are given analytically, and provide the optimal solution to \eqref{eq:valfunc}.
In practice, the gradient $\deriv$ is saved as look-up tables over a grid representing the state space of the relative system.

\subsection{Error Bound Guarantee via Value Function} \label{sec:proofs}
Prop. \ref{prop:nonconv} states the main theoretical result of this paper\footnote{The analogous infinite time horizon case is proven in the Appendix}: every level set of $\valfunc(\rstate, \tvar)$ is invariant under the following conditions:
\begin{enumerate}
  \item The tracking model applies the control in \eqref{eq:opt_ctrl_fin} which tries to track the planning model;
  \item The planning model applies the control in \eqref{eq:opt_dstb_fin} which tries to escape from the tracking model; \label{ln:plan}
  \item The tracking model experiences the worst-case disturbance in \eqref{eq:opt_dstb_fin} which tries to prevent successful tracking. \label{ln:dist}
\end{enumerate}

In practice, since the planning control and disturbance are \textit{a priori} unknown and are not directly controlled by the tracking model, conditions \ref{ln:plan} and \ref{ln:dist} may not hold. In this case, our theoretical results still hold; in fact, the absence of conditions \ref{ln:plan} and \ref{ln:dist} is advantageous to the tracking model and makes it ``easier'' to stay within its current level set of $\valfunc(\rstate, \tvar)$.
The smallest level set corresponding to the value $\underline\valfunc := \min_{\rstate} \valfunc(\rstate,\thor)$ can be interpreted as the smallest possible tracking error of the system.
The TEB is given by the set\footnote{In practice, since $\valfunc$ is obtained numerically, we set $\TEB(\tau) = \{\rstate: \valfunc(\rstate, \thor - \tau) \le \underline\valfunc + \epsilon\}$ for some suitably small $\epsilon>0$.}

\begin{align} \label{eq:TEB_fin}
\TEB(\tau) = \{\rstate: \valfunc(\rstate, \thor - \tau) \le \underline\valfunc\}.
\end{align}

Recall that we write the relative system state as $\rstate = (\estate, \astate)$, where $\estate,\astate$ are the error and auxiliary states.
Therefore, the TEB in the error state subspace is given by projecting away the auxiliary states $\astate$ in $\TEB(\tau)$:

  \begin{align} \label{eq:TEBp_fin}
  \TEB_\estate(\tau) = \{\estate: \exists \astate, \valfunc(\estate, \astate, \thor-\tau) \le \underline\valfunc\}
  \end{align}

This is the TEB that will be used in the online framework as shown in Fig. \ref{fig:fw_online}.
Within this bound the tracking model may use any controller, but on the boundary\footnote{Practical issues arising from sampled data control can be handled using methods such as \cite{Mitchell2012, Mitchell13, Dabadie2014} and are not the focus of our paper.} of this bound the tracking model must use the optimal tracking controller.
In general, the TEB is defined as a set in the error space, which allows the TEB to not only be in terms of position, but any state of the planning model such as velocity, as demonstrated in the example in Section \ref{sec:resultsMPC}.

We now formally state and prove the proposition.

\begin{prop}
  \label{prop:nonconv}
  \textbf{Finite time horizon guaranteed TEB.}
  Given $\tvar \in [0, \thor]$,

  \begin{subequations} \label{eq:fin_thor_prop}
      \begin{align}
      &\forall \tvar' \in [\tvar, \thor], ~ \rstate \in \TEB(\tvar) \Rightarrow \rtraj^*(\tvar'; \rstate, \tvar) \in \TEB(\tvar'), \text{ where} \label{eq:fin_thor_prop:statement}\\
       &\quad \rtraj^*(\tvar'; \rstate, \tvar) := \rtraj(\tvar'; \rstate, \tvar, \tctrl^*(\cdot), \pctrl^*(\cdot), \dstb^*(\cdot))), \label{eq:fin_thor_prop:here} \\
      &\quad
      \begin{aligned}
      &\tctrl^*(\cdot) = \arg \inf_{\tctrl(\cdot)\in\tcfset(t)}\big\{\\
      & \quad \max_{\tvar' \in [\tvar, \thor]} \errfunc(\rtraj(\tvar'; \rstate, \tvar, \tctrl(\cdot), \pctrl^*(\cdot), \dstb^*(\cdot))) \big\}, \label{eq:fin_thor_prop:ctrl}\\
      \end{aligned} \\
      &\quad
      \begin{aligned}
      & \pctrl^*(\cdot) := \gamma_\pstate^*[\tctrl](\cdot) = \arg \sup_{\gamma_{\pstate} \in \Gamma_\pstate(t)} \inf_{\tctrl(\cdot) \in \tcfset(\tvar)} \big\{ \\
      & \quad \max_{\tvar' \in [\tvar, \thor]} \errfunc(\rtraj(\tvar'; \rstate, \tvar, \tctrl(\cdot), \gamma_\pstate[\tctrl](\cdot), \dstb^*(\cdot))) \big\} \\
      \end{aligned} \\
      &\quad
      \begin{aligned}
      & \dstb^*(\cdot) = \arg \sup_{\gamma_{\dstb} \in \Gamma_\dstb(t)} \sup_{\gamma_{\pstate} \in \Gamma_\pstate(t)} \inf_{\tctrl(\cdot) \in \tcfset(t)} \big\{\\
      & \quad \max_{\tvar' \in [\tvar, \thor]} \errfunc(\rtraj(\tvar'; \rstate, \tvar, \tctrl(\cdot), \gamma_\pstate[\tctrl](\cdot), \gamma_\dstb[\tctrl](\cdot))) \big\}
      \end{aligned} \label{eq:fin_thor_prop:there}
      \end{align}
  \end{subequations}

\end{prop}

\textit{Proof:}
  We first show that given $\tvar \in [0, \thor]$,

  \begin{equation} \label{eq:vf_nondec}
    \forall \tvar' \in [\tvar, \thor], ~\valfunc(\rstate, \thor - \tvar) \ge \valfunc(\rtraj^*(\tvar'; \rstate, \tvar), \thor - \tvar')
  \end{equation}

  This follows from the definition of value function.

  \begin{subequations} \label{eq:fin_thor_steps}
    \begin{align}
      \valfunc(\rstate, \thor - \tvar) & = \max_{\tau \in [0, \thor-\tvar]} \errfunc(\rtraj^*(\tau; \rstate, 0)) \label{eq:fin_thor_steps:1} \\
      & = \max\{ \max_{\tau \in [0, \tvar'-\tvar]} \errfunc(\rtraj^*(\tau; \rstate, 0)), \nonumber\\
      &\qquad\qquad \max_{\tau \in [\tvar'-\tvar, \thor-\tvar]} \errfunc(\rtraj^*(\tau; \rstate, 0)) \} \label{eq:fin_thor_steps:2}\\
      & \ge \max_{\tau \in [\tvar'-\tvar, \thor-\tvar]} \errfunc(\rtraj^*(\tau; \rstate, 0)) \label{eq:fin_thor_steps:3}\\
      & = \max_{\tau \in [0, \thor-\tvar']} \errfunc(\rtraj^*(\tau; \rstate, \tvar-\tvar')) \label{eq:fin_thor_steps:4}\\
      & = \max_{\tau \in [0, \thor-\tvar']} \errfunc(\rtraj^*(\tau; \rtraj^*(0; \rstate, \tvar-\tvar'), 0)) \label{eq:fin_thor_steps:5}\\
      & = \max_{\tau \in [0, \thor-\tvar']} \errfunc(\rtraj^*(\tau; \rtraj^*(\tvar'; \rstate, \tvar), 0)) \label{eq:fin_thor_steps:6}\\
      & = \valfunc(\rtraj^*(\tvar'; \rstate, \tvar), \thor - \tvar') \label{eq:fin_thor_steps:7}
    \end{align}
  \end{subequations}

Explanation of steps:
\begin{itemize}
  \item \eqref{eq:fin_thor_steps:1}, \eqref{eq:fin_thor_steps:7}: by definition of value function, after shifting the time interval in \eqref{eq:fin_thor_prop:ctrl} to \eqref{eq:fin_thor_prop:there} from $[\tvar, \thor]$ to $[0, \thor-\tvar]$.
  \item \eqref{eq:fin_thor_steps:2}: rewriting $\max_{\tau \in [0, \thor-\tvar]}$ by splitting up the time interval $[0, \thor-\tvar]$ into $[0, \tvar'-\tvar]$ and $[\tvar'-\tvar, \thor-\tvar]$
  \item \eqref{eq:fin_thor_steps:3}: ignoring first argument of the outside $\max$ operator
  \item \eqref{eq:fin_thor_steps:4}: shifting time reference by $\tvar-\tvar'$, since dynamics are time-invariant
  \item \eqref{eq:fin_thor_steps:5}: splitting trajectory $\rtraj^*(\tau; \rstate, \tvar-\tvar')$ into two stages corresponding to time intervals $[\tvar-\tvar', 0]$ and $[0, \tau]$
  \item \eqref{eq:fin_thor_steps:6}: shifting time reference in $\rtraj^*(0; \rstate, \tvar-\tvar')$ by $\tvar'$, since dynamics are time-invariant
\end{itemize}

Now, we finish the proof as follows:

\begin{subequations} \label{eq:fin_hor}
  \begin{align}
  \rstate \in \TEB(\tvar) &\Leftrightarrow \valfunc(\rstate, \thor - \tvar) \le \underline\valfunc \\
  & \Rightarrow  \valfunc(\rtraj^*(\tvar'; \rstate, \tvar), \thor - \tvar') \le \underline\valfunc \label{eq:fin_hor:2}\\
  & \Leftrightarrow \rtraj^*(\tvar'; \rstate, \tvar) \in \TEB(\tvar'),
  \end{align}
\end{subequations}

\noindent where $\eqref{eq:vf_nondec}$ is used for the step in \eqref{eq:fin_hor:2}. \hfill $\blacksquare$

\subsection{Discussion}

\subsubsection{Worst-case assumptions} Prop. \ref{prop:nonconv} assumes that the planning control $\pctrl$ and disturbance $\dstb$ are optimally maximizing the value function $\valfunc$, and thereby increasing the size of the TEB $\TEB$.
Despite this, \eqref{eq:fin_thor_prop:statement} still holds.
In reality, $\pctrl$ and $\dstb$ do not behave in a worst-case fashion, and it is often the case that when $\tvar' \ge \tvar$, we have $\rstate \in \TEB(\tvar) \Rightarrow \rtraj(\tvar'; \rstate, \tvar) \in \TEB(\tau)$ for some $\tau \le \tvar'$.
Thus, one can ``take advantage'' of the suboptimality of $\pctrl$ and $\dstb$ by finding the earliest $\tau$ such that $\rtraj(\tvar'; \rstate, \tvar) \in \TEB(\tau)$ in order to have the tighter TEB over a longer time-horizon.

\subsubsection{Relationship to reachable sets} Prop. \ref{prop:nonconv} is similar to well-known results in differential game theory with a different cost function \cite{Akametalu2014}, and has been utilized in the context of interpreting the subzero level set of $\valfunc$ as a backward reachable set for tasks such as collision avoidance \cite{Mitchell05}. In this work we do not assign special meaning to any particular level set, and instead consider all level sets at the same time. This allows us to effectively solve many simultaneous reachability problems in a single computation, removing the need to check whether resulting invariant sets are empty, as was done in \cite{Bansal2017}.

\subsubsection{Relationship to control-Lyapunov functions} One interpretation of \eqref{eq:TEBp_fin} is that $\valfunc(\rstate, \thor - \tvar)$ is a control-Lyapunov function for the relative dynamics between the tracking model and the planning model, and any level set of $\valfunc(\rstate, \thor - \tvar)$ is invariant.
It should be noted that computing control-Lyapunov functions for general nonlinear systems is difficult.
In addition, the relative system trajectories are guaranteed to remain within the initial level set despite the worst-case disturbance.
In the absence of any information about, for example, the intent of the planning system, such a worst-case assumption is needed.

\subsubsection{Guaranteed stability under switching controllers}
Suppose that the relative system state $r(t)$ is in the interior of some $\alpha$-sublevel set of $V$, $\{r:V(r,T-t)\le \alpha\}$, for any $\alpha,t$.
Then, As long as the optimal tracking control is applied whenever $V(r(t'),T-t')=\alpha$ (at any time $t'\ge t$), the relative state $r(s)$ would remain inside $\{r:V(r,T-s)\le \alpha\}$ for all $s\ge t'$.
This means that when $V(r(t),t)< \alpha$, any controller can be used without affecting the stability in the sense of the relative system remaining inside some $\alpha$ sublevel set of $V$.
Switching to the optimal controller whenever $V(r(t'),t' )=\alpha$ guarantees stability regardless of controller switching.

\example{In this example we have computed a converged value function $\valfunc_\infty$.
The corresponding TEB can be found using \eqref{eq:TEB_inf} in the Appendix.
We can similarly find the TEB projected onto the planning states using \eqref{eq:TEBp_inf}.
The minimum of the value function was approximately $\underline\valfunc = 0.004$, and the size of the TEB in $(x_r, y_r)$ space is approximately $0.065$. The converged value function and TEB can be seen in Fig. \ref{fig:vf_TEB:5D3D}.
The corresponding optimal tracking controller is obtained by plugging the gradients of our converged value function and our relative system dynamics into (\ref{eq:opt_ctrl_inf}).
}
\section{Online Computation \label{sec:online}}
Algorithm \ref{alg:algOnline} describes the online computation.
Lines \ref{ln:gStart} to \ref{ln:gEnd} indicate that the value function $\valfunc(\rstate,\tvar'')$, the gradient $\deriv$ from which the optimal tracking controller is obtained, as well as the TEB sets $\TEB,\TEB_\estate$ are given from offline precomputation.
Lines \ref{ln:Istart}-\ref{ln:Iend} initialize the computation by setting the planning and tracking model states such that the relative system state is inside the TEB $\TEB$.

\begin{algorithm}
	\caption{Online Trajectory Planning}
	\label{alg:algOnline}
	\begin{algorithmic}[1]
		\STATE \textbf{Given}: \label{ln:gStart}
		\STATE $\valfunc(\rstate, \tvar''), \tvar'' \in [0, \thor]$ and gradient $\nabla \valfunc(\rstate, \tvar'')$
 		\STATE $\TEB(\tvar'), \tvar' \in [0,\thor]$ from \eqref{eq:TEB_fin}, and $\TEB_\estate$ from \eqref{eq:TEBp_fin} \label{ln:gEnd}
    \STATE \textbf{Initialization}: \label{ln:Istart}
		\STATE Choose $\pstate, \tstate$ such that $\rstate \in \TEB(0)$
    \STATE Set initial time: $\tvar \leftarrow 0$. \label{ln:Iend}
		\WHILE{Planning goal is not reached OR planning horizon is exceeded}
		\STATE \textbf{TEB Block}: \label{ln:obsStart}
    \STATE Look for the smallest $\tau$ such that $\rstate \in \TEB(\tau)$ \label{ln:infSkip}
		\STATE $\constrAug(\tvar + \tvar') \leftarrow \constrSense \ominus \TEB_\estate(\tau + \tvar')$ \label{ln:obsEnd}

		\STATE \textbf{Path Planning Block}:\label{ln:plannerStart}
		\STATE $\pstate_\text{next} \leftarrow \plannerfunc(\pstate, \constrAug)$\label{ln:plannerEnd}

		\STATE \textbf{Hybrid Tracking Controller Block}:\label{ln:controllerStart}
		\STATE $\rstate_\text{next} \leftarrow \rtrans(\tstate,\pstate)(\tstate - \ptmat\pstate_\text{next})$

		\IF{$\rstate_\text{next}$ is on boundary $\TEB_\estate(\tvar)$}
		\STATE {use optimal tracking controller: $\tctrl \leftarrow \tctrl^*$ in \eqref{eq:opt_ctrl_fin}}
		\ELSE \STATE{use performance controller: }
          \STATE{$\tctrl \leftarrow$ desired controller} \ENDIF \label{ln:controllerEnd}

		\STATE \textbf{Tracking Model Block}: \label{ln:trackingStart}
		\STATE apply control $\tctrl$ to vehicle for a time step of $\dt$
    \STATE the control $\tctrl$ and disturbance $\dstb$ bring the system to a new state $\tstate$ according to \eqref{eq:tdyn} \label{ln:trackingEnd}

		\STATE \textbf{Planning Model Block}:\label{ln:planningStart}
		\STATE update planning state, $\pstate \leftarrow \pstate_\text{next}$, from Line \ref{ln:plannerEnd}
		\STATE check if $\pstate$ is at planning goal \label{ln:planningEnd}
    \STATE \textbf{Update time}:
    \STATE $\tvar \leftarrow \tvar + \Delta \tvar$
		\ENDWHILE
	\end{algorithmic}
\end{algorithm}

The TEB block is shown on lines \ref{ln:obsStart}-\ref{ln:obsEnd}.
The sensor detects obstacles, or in general constraints, $\constrSense(\cdot)$ within the sensing region around the vehicle.
If the user allows the planning model to instantaneously stop, then for the static environments explored in this paper the sensing region must be large enough to sense any obstacles within one TEB of the planning algorithm.
Thus, the set representing the minimum allowable sensing region\footnote{$T\rightarrow\infty$ for the infinite time horizon case.} is $\senseDist = \{\TEB_\estate(\thor)\} \bigoplus FRS(\delta \tvar)$, where $FRS(\delta \tvar)$ is the forward reachable set of the planning model for one time step of planning (i.e. the largest step in space that the planning algorithm can make in one time step).
When using sensors that perceive some fixed radius in position in all directions, the required radius is simply the maximum distance of the TEB summed with the maximum distance the planning algorithm can cover in one time step.

Note that for time-varying TEBs with long time horizons this sensing requirement can be fairly restrictive depending on the maximum size of the time-varying TEB.
If the planning model is not allowed to stop instantaneously, recursive safety can be ensured by methods such as \cite{bajcsy2019efficient, fridovich2018safe}.
FaSTrack applied to dynamic environments with humans is explored in \cite{fisac2018probabilistically,  fridovich2019confidence}, and is paired with work on sequential trajectory tracking \cite{chen2018robust} to handle multi-human, multi-robot environments \cite{bajcsy2018scalable}.

Constraints are defined in the state space of the planning model, and therefore can represent constraints not only in position but also in, for example, velocity or angular space. One can either augment the constraints by the TEB, or augment the planning algorithm by the TEB.
Augmenting either planning algorithms or constraints by some buffer is common practice in motion planning.  The decision on which to augment falls to the user based on the planning method used.  Augmenting the planning algorithm requires computing the intersection of sets between the TEB and the constraints (as done in \cite{fisac2018probabilistically, bajcsy2018scalable}). Augmenting the constraints instead requires using the Minkowski difference, denoted ``$\ominus$.''  If the TEB is a complicated shape for which computing the Minkowski difference is difficult, one can reduce computational speed by simply expanding the constraints by the maximum distance of the TEB in each dimension (this will result in a more conservative approximation of the unsafe space).

The path planning block (lines \ref{ln:plannerStart}-\ref{ln:plannerEnd}) takes in the planning model state $\pstate$ and the augmented constraints $\constrAug$, and outputs the next state of the planning model $\pstate_\text{next}$ through the function $\plannerfunc(\cdot, \cdot)$.
As mentioned, FaSTrack is agnostic to the planning algorithm used, so we assume that $\plannerfunc(\cdot, \cdot)$ has been provided.
The hybrid tracking controller block (lines \ref{ln:controllerStart}-\ref{ln:controllerEnd}) first computes the updated relative system state $\rstate_\text{next}$.
If the $\rstate_\text{next}$ is on the boundary of the TEB $\TEB_\estate(0)$, the optimal tracking controller given in \eqref{eq:opt_ctrl_inf} must be used to remain within the TEB.
If the relative system state is not on the tracking boundary, a performance controller may be used. For the example in Section \ref{sec:results} the safety and performance controllers are identical, but in general this performance controller can suit the needs of the individual applications.

The control $\tctrl^*$ is then applied to the physical system in the tracking block (lines \ref{ln:trackingStart}-\ref{ln:trackingEnd}) for a time period of $\dt$.
The next state is denoted $\tstate_\text{next}$.
Finally, the planning model state is updated to $\pstate_\text{next}$ in the planning model block (lines \ref{ln:planningStart}-\ref{ln:planningEnd}).
We repeat this process until the planning goal has been reached.
\section{Numerical examples} \label{sec:results}

In this section, we demonstrate the FaSTrack framework in three numerical simulation examples that respectively represent dynamic programming-based, sampling-based, and optimization-based planning algorithms: (1) a 5D car tracking a 3D car model with the FSM planning algorithm, (2) a 10D quadrotor tracking a single integrator model with the RRT planning algorithm, and (3) an 8D quadrotor tracking a double integrator model with the MPC planning algorithm.
In each example, obstacles in the environment are \textit{a priori} unknown, and are revealed to the vehicle when they are ``sensed," i.e. come within the minimum allowable sensing distance.
Whenever the obstacle map is updated, the planning algorithm replans a trajectory in real time.
In this paper, the details of sensing are kept as simple as possible; we aim to only demonstrate our framework for real-time guaranteed safe planning and replanning.
In general, any other planning algorithm can be used for planning in unknown environments, as long as planning and replanning can be done in real time.

For each example, we first describe the tracking and planning models.
Next, we present the relative dynamics as well as the precomputation results.
Afterwards, we briefly describe the planning algorithm and how obstacles are sensed by the vehicle.
Finally, we show trajectory simulation results.

\subsection{\textbf{Running Example:} 5D car-3D car example with FSM \label{sec:reach_planner}}

For our first example, we continue our running example of the 5D tracking model and 3D planning model of an autonomous car. We will demonstrate the combination of fast planning and provably robust tracking by combining the fast sweeping method (FSM) \cite{Takei2013} with our computed TEB.
FSM is an efficient optimal control-based planning algorithm for car-like systems, and provides numerically convergent globally optimal trajectory in real time.
In this example, we use FSM to perform real-time planning for the 3D kinematic car model, whose trajectory is tracked by the 5D car model.

\subsubsection{Offline computation}

\begin{figure}
	\includegraphics[width=\columnwidth]{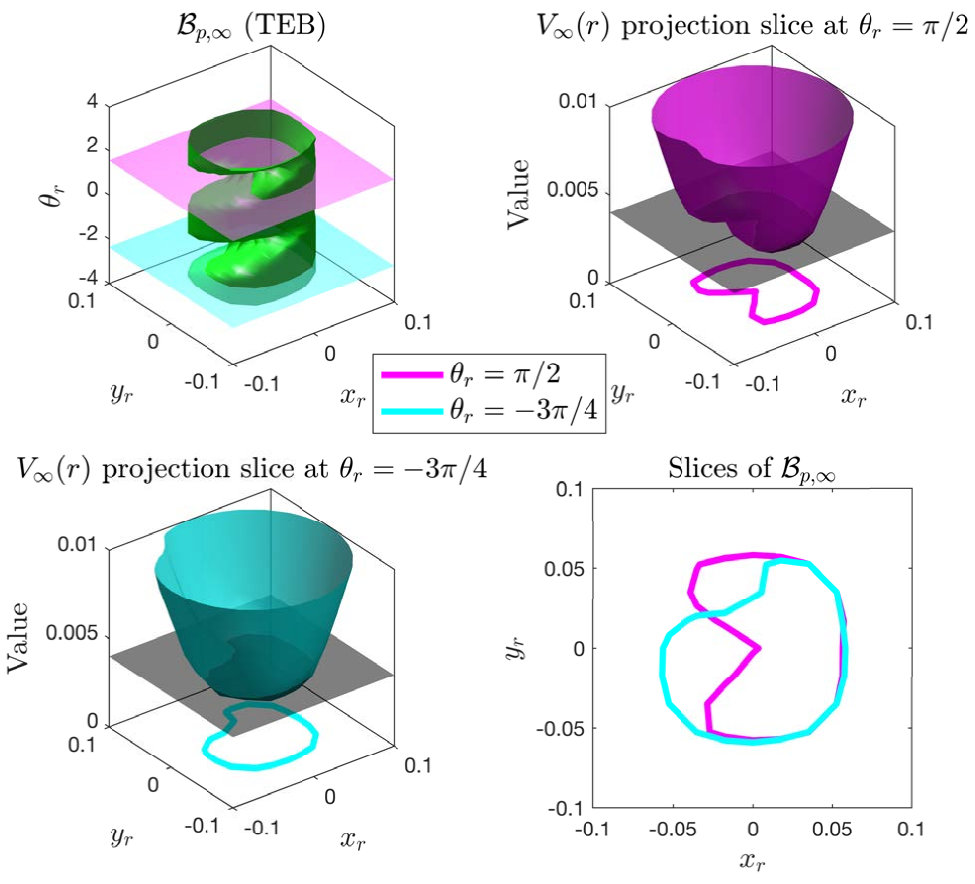}
	\caption{Infinite time horizon TEB (top left), two slices of the value function at $\theta_r = \pi/2, -3\pi/4$ (top right, bottom left), and corresponding TEB slices (bottom right) for the running example (5D car tracking 3D car) introduced in Section \ref{sec:reach_planner}.}

	\label{fig:vf_TEB:5D3D}
\end{figure}

As stated in Sec. \ref{sec:precomp}, we computed the converged value function $\valfunc_\infty(\rstate)$, which is shown in Fig. \ref{fig:vf_TEB:5D3D}, with $\underline\valfunc = 0.004$, and $\TEB_{\pstate, \infty} = 0.065$.
The top left plot of Fig. \ref{fig:vf_TEB:5D3D} shows the TEB $\TEB_{\pstate, \infty}$ in green.
The tracking model must apply the optimal control when it is on the green boundary.
The cross-sectional area of the TEB is the largest at $\theta_r = 0,\pi$, because at these $\theta_r$ values the 5D car model is either aligned with or opposite to the 3D car model.
Since the 5D car is able to move both forward and backward, these two alignments make tracking the easiest.
For the same reasoning, the cross sectional area is the smallest at $\theta_r = -\pi/2, \pi/2,$ etc.

The magenta and cyan planes indicate slices of the TEB at $\theta_r = \pi/2, -3\pi/4$, respectively.
With these $\theta_r$ values fixed, corresponding projections of the value function onto $(x_r, y_r)$ space are shown in the top right and bottom left plots.
Here, $\underline\valfunc$ is shown as the gray plane, with the intersection of the gray plane and the value function projection shown by the curve in the $0$-level plane.
These curves are slices of $\TEB_{\pstate, \infty}$ at the $\theta_r = \pi/2, -3\pi/4$ levels.

Computation was done on a desktop computer with an Intel Core i7 5820K processor, on a $31\times31\times45\times27\times47$ grid, and took approximately 23 hours and required approximately 2 GB of RAM using a C++ implementation of level set methods for solving \eqref{eq:HJVI}.
A 5D computation is at the limit of computational tractability using the HJ method.
Fortunately, FaSTrack is modular with respect to the method for computating the TEB, and we are exploring techniques for computing TEBs for higher-dimensional systems through sum-of-squares optimization \cite{SinghChenEtAl2018} and approximate dynamic programming \cite{royo2018classification}.

\subsubsection{Online sensing and planning}
\begin{figure}
  \centering
  \begin{subfigure}[t]{0.49\columnwidth}
    \includegraphics[width=\columnwidth]{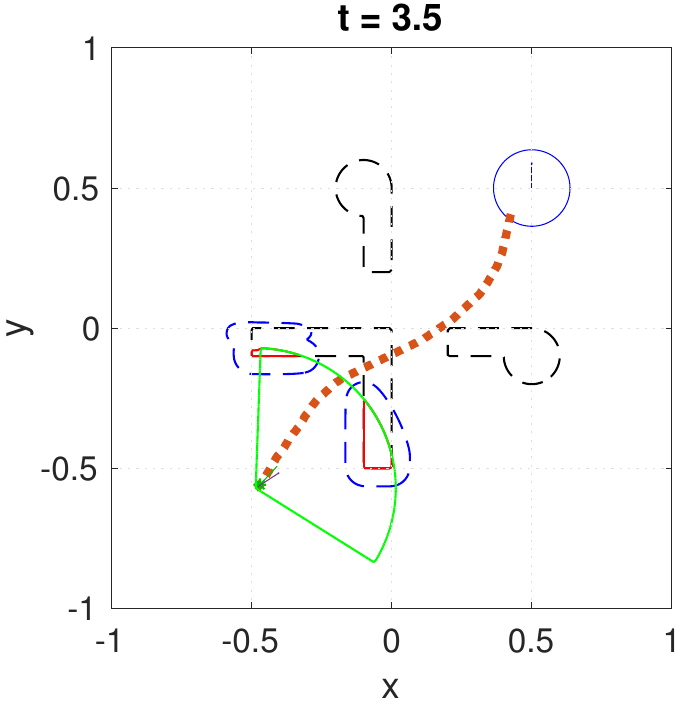}
  \end{subfigure}
  \begin{subfigure}[t]{0.49\columnwidth}
    \includegraphics[width=\columnwidth]{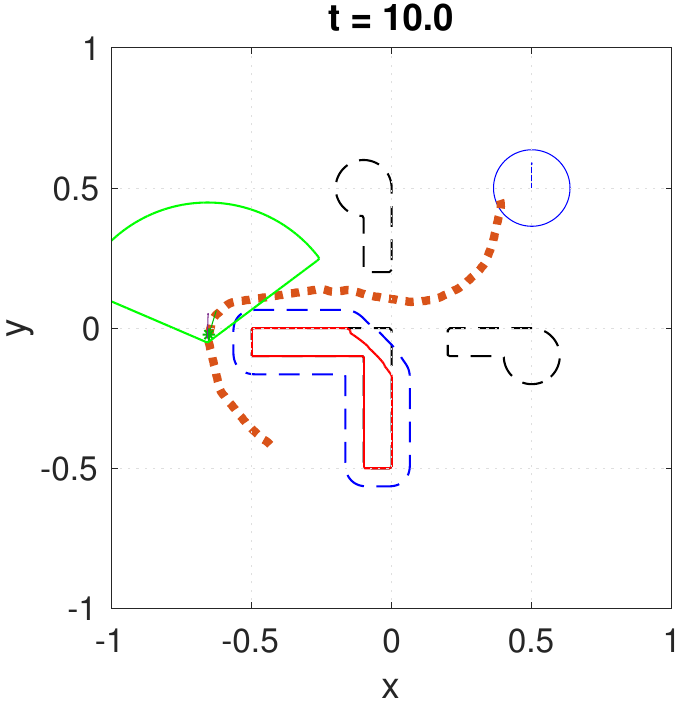}
  \end{subfigure}

  \begin{subfigure}[t]{0.49\columnwidth}
    \includegraphics[width=\columnwidth]{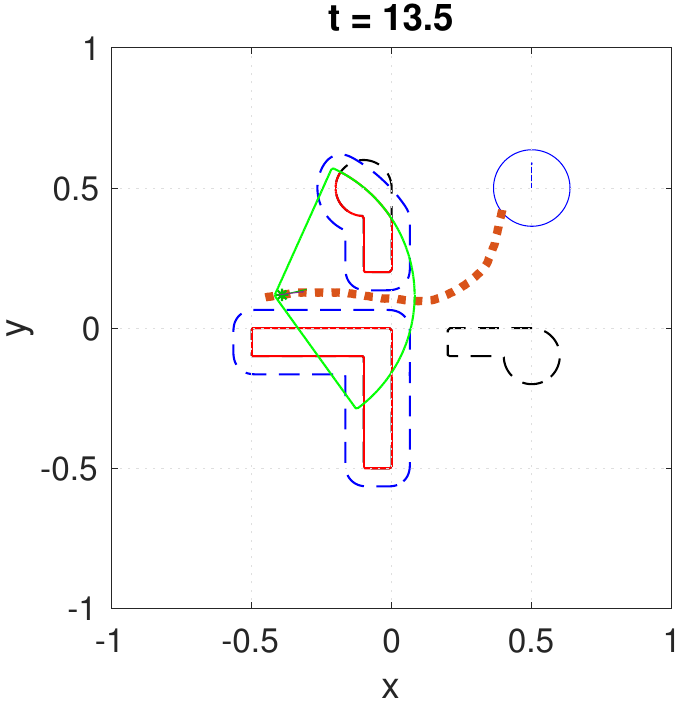}
  \end{subfigure}
  \begin{subfigure}[t]{0.49\columnwidth}
    \includegraphics[width=\columnwidth]{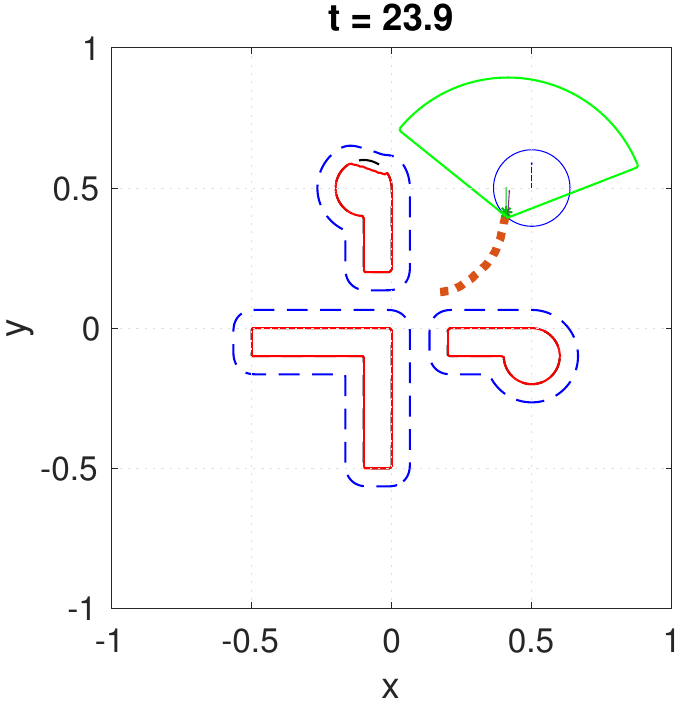}
  \end{subfigure}
  \caption{Simulation of the 5D-3D example. As the vehicle with 5D car dynamics senses new obstacles in the sensing region (light green), the 3D model replans trajectories, which are robustly tracked by the 5D system. Augmentation of the constraints resulting from the obstacles ensures safety of the 5D system using the optimal tracking controller.}
  \label{fig:5D3Dsim}
\end{figure}

The simulation showing the combination of tracking and planning is shown in Fig. \ref{fig:5D3Dsim}.
The goal of the system, the 5D car, is to reach the blue circle at $(0.5, 0.5)$ with a heading that is within $\pi/6$ of the direction indicated by the arrow inside the blue circle, $\pi/2$.
Three initially unknown obstacles, whose boundaries are shown in dotted black, make up the constraints $\constr$.

While planning a trajectory to the goal, the car also senses obstacles.
For this example, we chose a simple virtual sensor that reveals obstacles within a range of $0.5$ and in front of the vehicle within an angle of $\pi/6$, depicted as the light green fan.
When a portion of the unknown obstacles is within this region, that portion is made known to the vehicle, and is shown in red.
These make up the sensed constraints $\constrSense$.
To ensure that the 5D car does not collide with the obstacles despite error in tracking, planning is done with respect to augmented constraints $\constrAug$, shown in dashed blue.

Given the current planning constraints $\constrAug$, the planning algorithm uses the 3D planning model to generate a trajectory, in real time using FSM, towards the goal.
This plan is shown in dotted red.
The 5D system robustly tracks the 3D system within the TEB in Fig. \ref{fig:vf_TEB:5D3D}.
Four time snapshots of the simulation are shown in Fig. \ref{fig:5D3Dsim}.
In the top left subplot, the system has sensed only a very small portion of the obstacles, and hence plans a trajectory through an unknown obstacle to the target.
However, while tracking this initial trajectory, more of the L-shaped obstacle is made known to the system, and therefore the system plans around this obstacle, as shown in the top right subplot.
The bottom subplots show the system navigating through sensed obstacles and reaching the goal at $t=23.9$ s.

\begin{figure}
  \includegraphics[width=\columnwidth]{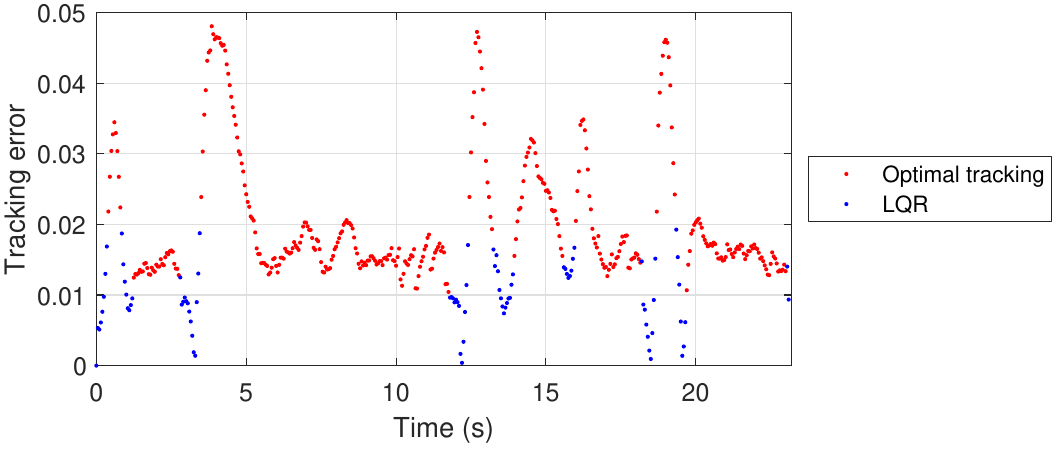}
  \caption{Tracking error bound over time for the 5D-3D example. The red dots indicate that the optimal tracking controller is used, while the blue dots indicate that an LQR controller for the linearized system is used. The hybrid controller switches from LQR to the optimal tracking controller whenever the error exceeds 0.02. The tracking error is always well below the predicted TEB of 0.065.}

  \label{fig:P5D_Dubins_tracking_error}
\end{figure}

As explained in Fig. \ref{fig:hybrid_ctrl}, when the tracking error is relatively large, the autonomous system uses the optimal tracking controller given by \eqref{eq:opt_ctrl_inf}; otherwise, it uses a performance controller.
In this simulation, we used a simple LQR controller on the linearized system when the tracking error is less than a quarter of the size of the TEB.
In general, this switching condition is user-defined.
The tracking error over time is shown in Fig. \ref{fig:P5D_Dubins_tracking_error}.
The red dots indicate the time points at which the optimal tracking controller in \eqref{eq:opt_ctrl_inf} is used, and the blue dots indicate the time points at which the LQR controller is used.
One can see that when the optimal tracking controller is used, the error stays below 0.05, well below the predicted TEB of 0.065, since the planning control and the disturbances are not being adversarial.
The disturbance was chosen to be uniformly random within the chosen bounds.

The simulation was done in MATLAB on a desktop computer with an Intel Core i7 2600K CPU.
Time was discretized in increments of $0.067$ seconds, (15 Hz).
Averaged over the duration of the simulation, planning with FSM took approximately $66$ ms per iteration, and obtaining the tracking control from \eqref{eq:opt_ctrl_inf} took approximately $2$ ms per iteration.

\subsection{10D quadrotor-3D single integrator example with RRT\label{sec:resultsRRT}}

Our second example involves a 10D near-hover quadrotor \cite{Bouffard12} as tracking model and a single integrator in 3D space as planning model.
Planning is done using RRT, a well-known sampling-based planning algorithm that quickly produces geometric paths from a starting position to a goal position \cite{Kuffner2000,Kavraki1996}.
Paths given by the RRT planning algorithm are converted to time-stamped trajectories by placing a maximum velocity in each dimension along the generated geometric paths.

The dynamics of tracking model and of the 3D single integrator is as follows:

\begin{equation}
\label{eq:Quad10D_dyn}
\begin{bmatrix}
\dot{x}\\
\dot{v_x}\\
\dot{\theta_x}\\
\dot\omega_x\\
\dot{y}\\
\dot{v_y}\\
\dot{\theta_y}\\
\dot\omega_y\\
\dot{z}\\
\dot{v_z}
\end{bmatrix}
=
\begin{bmatrix}
v_x + d_x\\
g \tan \theta_x\\
-d_1 \theta_x + \omega_x\\
-d_0 \theta_x + n_0 a_x\\
v_y + d_y\\
g \tan \theta_y\\
-d_1 \theta_y + \omega_y\\
-d_0 \theta_y + n_0 a_y\\
v_z + d_z\\
k_T a_z - g
\end{bmatrix}, \quad
\begin{bmatrix}
\dot{\hat x}\\
\dot{\hat y}\\
\dot{\hat z}\\
\end{bmatrix} =
\begin{bmatrix}
\hat v_x \\
\hat v_y \\
\hat v_z
\end{bmatrix},
\end{equation}
\noindent where quadrotor states $(x, y, z)$ denote the position, $(v_x, v_y, v_z)$ denote the velocity, $(\theta_x, \theta_y)$ denote the pitch and roll, and $(\omega_x, \omega_y)$ denote the pitch and roll rates.
The controls of the 10D system are $(u_x, u_y, u_z)$, where $u_x$ and $u_y$ represent the desired pitch and roll angle, and $u_z$ represents the vertical thrust.

The 3D system controls are $(\hat v_x, \hat v_y, \hat v_z)$, and represent the velocity in each positional dimension.
The disturbances in the 10D system $(\dstb_x, \dstb_y, \dstb_z)$ are caused by wind, which acts on the velocity in each dimension.
The model parameters are chosen to be $d_0=10$, $d_1=8$, $n_0=10$, $k_T=0.91$, $g=9.81$, $|u_x|, |u_y| \le \pi/9$, $u_z \in [0, 1.5g]$, $|\hat v_x|, |\hat v_y|, |\hat v_z| \le 0.5$.
The disturbance bounds were chosen to be $|d_x|, |d_y|, |d_z| \le 0.1$.

\subsubsection{Offline computation}
We define the relative system states to consist of the error states, or relative position $(x_r, y_r, z_r)$, concatenated with the rest of the state variables of the 10D quadrotor model.
Defining $\rtrans = \mathbf I_{10}$ and

\begin{equation*}
\ptmat =
\begin{bmatrix}
  \begin{bmatrix} 1 \\ \mathbf 0_{3 \times 1} \end{bmatrix}
    & \mathbf 0_{4\times 1}
    & \mathbf 0_{4\times 1} \\
  \mathbf 0_{4\times 1}
    & \begin{bmatrix} 1 \\ \mathbf 0_{3 \times 1} \end{bmatrix}
    &  \mathbf 0_{4\times 1} \\
  \mathbf 0_{2\times 1}
    & \mathbf 0_{2\times 1}
    & \begin{bmatrix} 1 \\ 0 \end{bmatrix}
\end{bmatrix},
\end{equation*}

\noindent we obtain the following relative system dynamics:

\begin{equation}
\label{eq:Quad10DRel_dyn}
\begin{bmatrix}
\dot x_r\\
\dot v_x\\
\dot \theta_x\\
\dot\omega_x\\
\dot y_r\\
\dot v_y\\
\dot \theta_y\\
\dot\omega_y\\
\dot z_r\\
\dot v_z
\end{bmatrix} =
\begin{bmatrix}
v_x - \hat v_x + d_x\\
g \tan \theta_x\\
-d_1 \theta_x + \omega_x\\
-d_0 \theta_x + n_0 u_x\\
v_y - \hat v_y + d_y\\
g \tan \theta_y\\
-d_1 \theta_y + \omega_y\\
-d_0 \theta_y + n_0 u_y\\
v_z - \hat v_z + d_z\\
k_T u_z - g
\end{bmatrix}.
\end{equation}

The relative system dynamics given in \eqref{eq:Quad10DRel_dyn} is decomposable into three independent subsystems involving the sets of variables $(x_r, v_x, \theta_x, \omega_x)$, $(x_y, v_y, \theta_y, \omega_y)$, $(z_r, v_z)$, allowing us to choose the error function to be also in the decomposable form of $\errfunc(\rstate) = \max(x_r^2, y_r^2, z_r^2)$, so that we can solve \eqref{eq:HJVI} tractably since each subsystem is at most 4D \cite{Chen2016DecouplingJournal}.

\begin{figure}
  \includegraphics[width=\columnwidth]{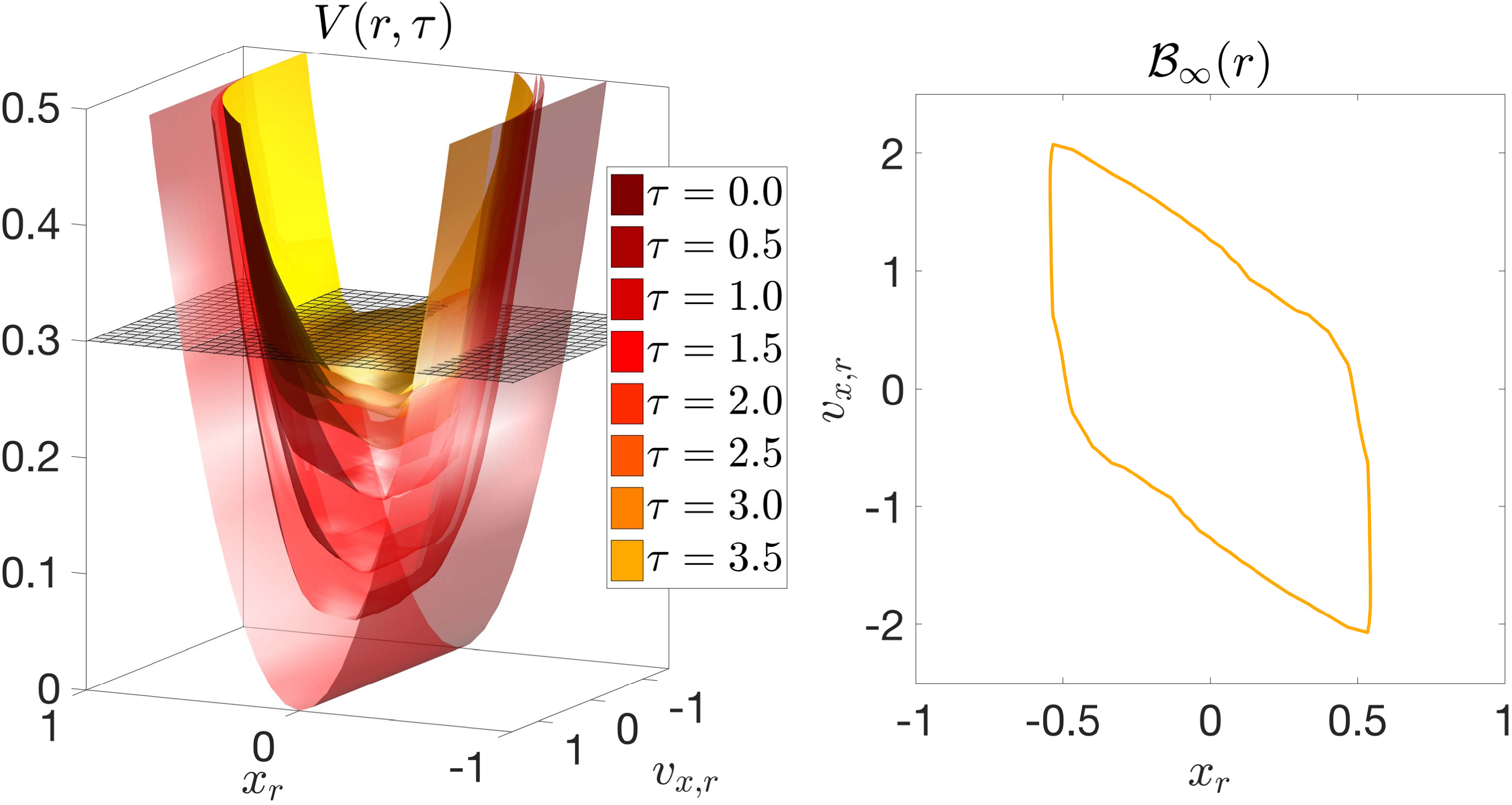}
  \caption{Left: snapshots in time of the value function $\valfunc(r,\tau)$ shown over dimensions $x_r$ and $v_{x,r}$.  Snapshots are from $\tau = 0$ s (transparant dark red surface on bottom) to convergence at $\tau = 3.5$ s (sollid yellow surface on top). Right: 2D slice at $\valfunc_\infty(\rstate)=0.3$ (corresponding to gray slice on the left). This is the infinite horizon TEB, $\TEB_\infty(\rstate)$ in the $x_r$ and $v_{x,r}$ dimensions.}
  \label{fig:valfuncRRT}
\end{figure}

The left subplot of Fig. \ref{fig:valfuncRRT} shows the projection of the value function $\valfunc$ onto the $(x_r, v_x)$ space resulting from solving \eqref{eq:HJVI} over an increasingly long time horizon.
Starting from $\tau=0$, we have that $\errfunc(\rstate) = \valfunc(\rstate, 0)$.
As $\tau$ increases, the value function evolves according to \eqref{eq:HJVI}, and eventually converges when $\tau$ reaches $3.5$.
This implies that $\valfunc_\infty(\rstate) = \valfunc(\rstate, \tau=3.5)$, since we would still obtain the same function even if we let $\tau$ approach infinity.
The horizontal plane shows $\underline\valfunc = 0.3$, which corresponds to a TEB of approximately 0.9.

The right subplot of Fig. \ref{fig:valfuncRRT} shows the $\underline\valfunc = 0.3$ level set of value function projection, which is the projection onto the $(x_r, v_x)$ space, of the TEB $\TEB_{\estate, \infty}$.
The range of $x_r$ provides the TEB used for the planning algorithm, $\TEB_{\pstate, \infty}$.
The value function and TEB in the $(y_r, v_y, \theta_y, \omega_y)$ and $(z_r, v_z)$ spaces are combined to form the 10D TEB, which is projected down to the 3D positional space.
For conciseness, these value functions are not shown; however, one can see the resulting TEB in Fig. \ref{fig:simRRT} and \ref{fig:simRRT_combined} as the translucent blue box.

Offline computations were done on a laptop with an Intel Core i7 4702HQ CPU using a MATLAB implementation of level set methods \cite{Mitchell07c} used for solving \eqref{eq:HJVI}.
The 4D computations were done on a $61\times 61 \times 41 \times 41$ grid, took approximately 12 hours, and required approximately 300 MB of RAM.
The 2D computation in the $(z_r, v_z)$ space was done on a $101 \times 101$ grid, took approximately 15 seconds, and required negligible RAM.

\subsubsection{Online sensing and planning}
\begin{figure}
  \includegraphics[width=\columnwidth]{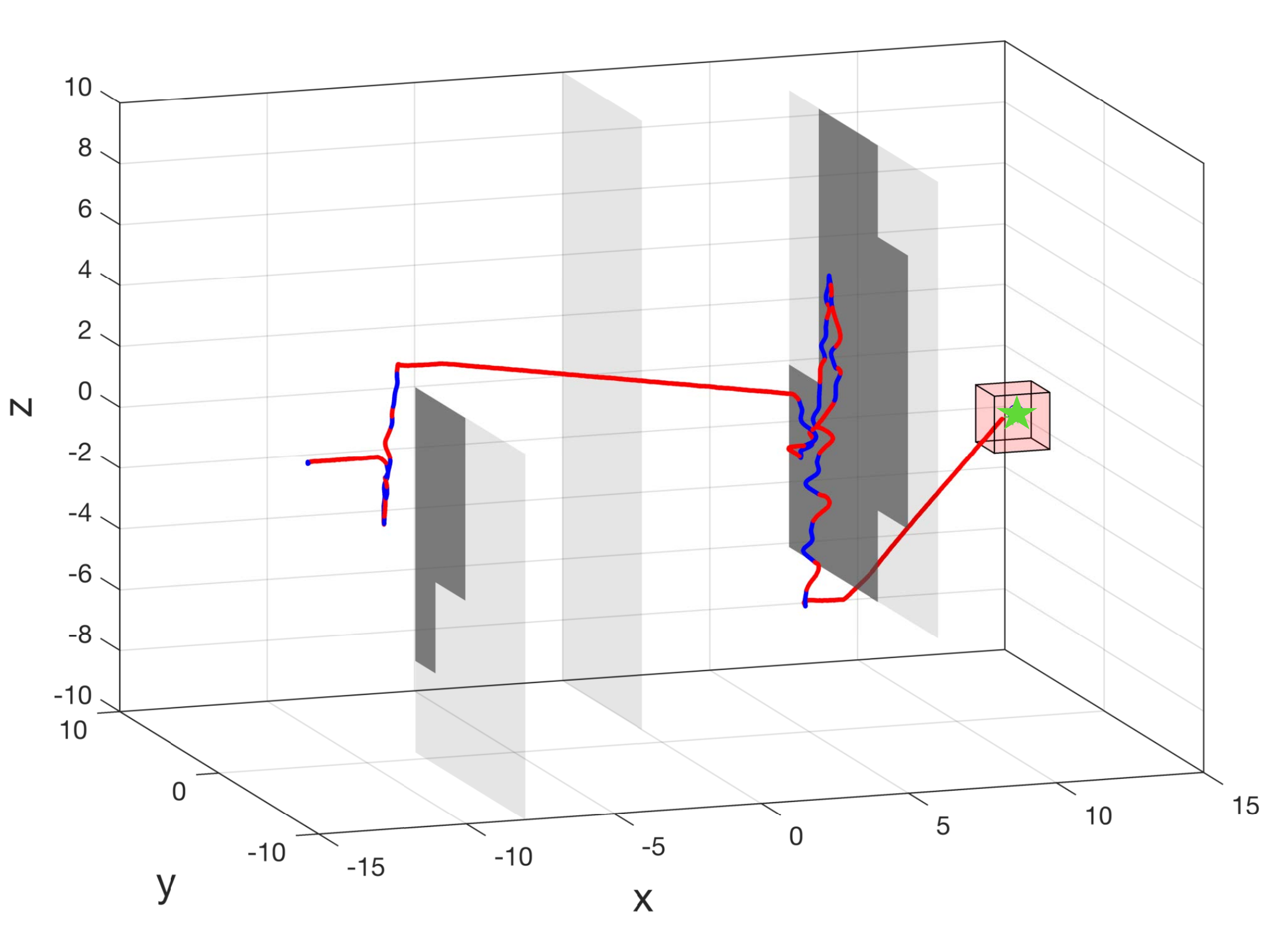}
  \caption{Simulation of the 10D quadrotor tracking a 3D single integrator (position shown as green star inside translucent red box). The dimensions $x,y,z$ represent the length, width, and height of the absolute state space.  The system senses initially unknown obstacles (light gray), which are revealed (revealed parts shown in dark gray) as the system approaches them. An LQR controller was used for tracking along the blue portions of the trajectory, and the optimal tracking controller was used along the red portions. Replanning is done in real time by RRT when new obstacles are sensed. The TEB is shown as the translucent red box, and is the set of positions that the 10D quadrotor is guaranteed to remain within.}
  \label{fig:simRRT}
\end{figure}

\begin{figure}
  \includegraphics[width=\columnwidth]{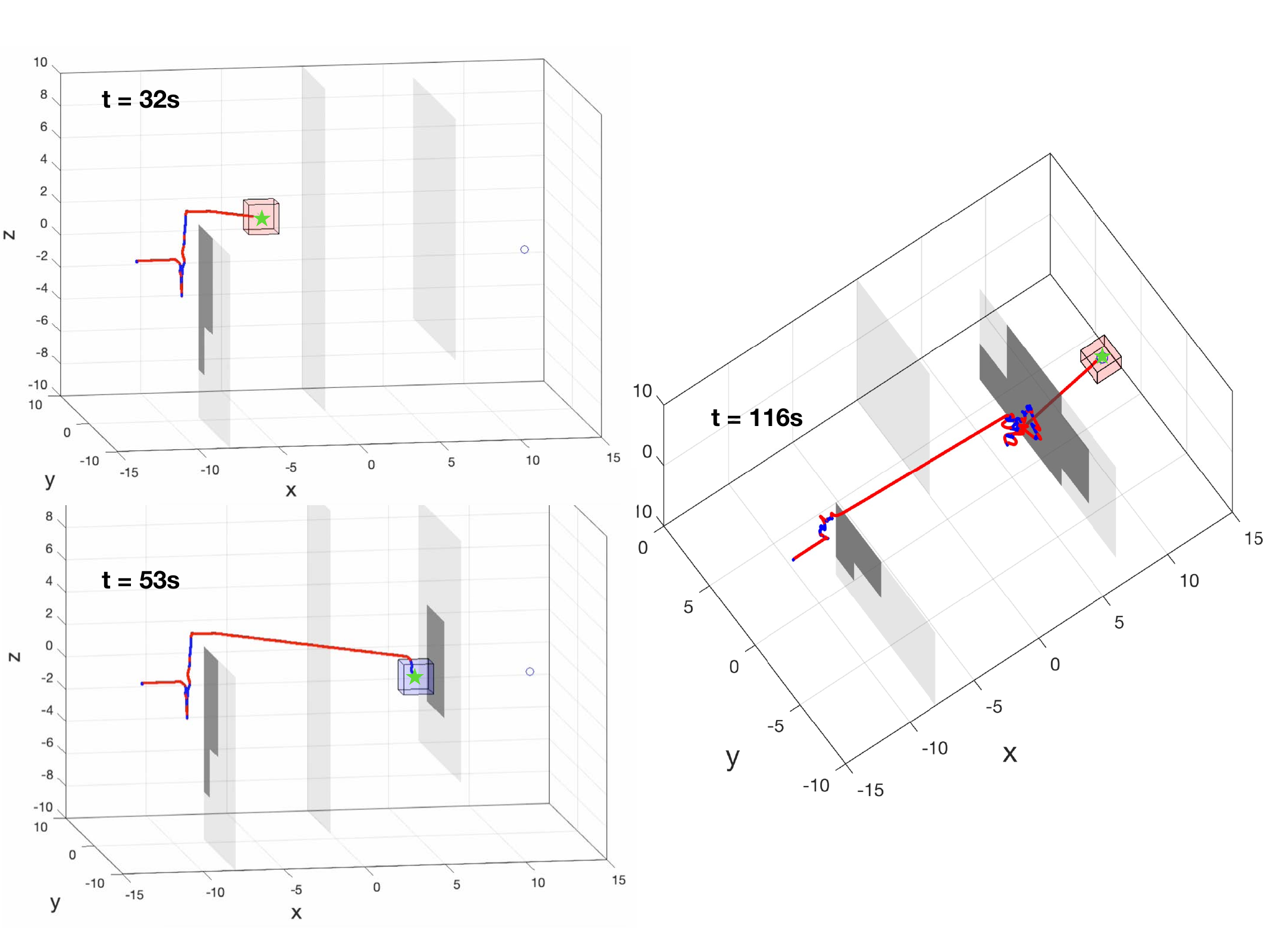}
  \caption{Three time snapshots of the simulation in Fig. \ref{fig:simRRT}. The full simulation can be seen at https://youtu.be/fR64\_LMdieA.}
  \label{fig:simRRT_combined}
\end{figure}

The simulation involving the 10D quadrotor model tracking the 3D single integrator is shown in Fig. \ref{fig:simRRT} and \ref{fig:simRRT_combined}.
Here, the system aims to start at $(x,y,z) = (-12, 0, 0)$ and reach $(12, 0, 0)$.
To best test our method, we allow the wind disturbance to act adversarially, resulting in worst-case wind conditions.
Three rectangular obstacles, which make up the constraints $\constr$, are present and initially unknown.
Before the obstacles are sensed by the system, they are shown in light gray.  As the 10D quadrotor senses obstacles (when they are within 1.5 units from the quadrotor), the portion of the obstacles that is within sensing distance is revealed and shown in dark gray.
To demonstrate the flexibility of augmenting by the TEB, in this example we show the TEB augmenting the planning algorithm rather than the obstacles.
Whenever new obstacles are revealed, the planning algorithm replans a trajectory to the goal while avoiding the augmented constraint set $\constrAug$. This happens in real time using RRT.

Fig. \ref{fig:simRRT} shows the entire trajectory, with the end of the trajectory being close to the goal position.
The planning model state is shown as a small green star, and the translucent red box around it depicts the TEB: the tracking model position is guaranteed to reside within this box.
Therefore, as long as the planning model plans in a way such that the TEB does not intersect with the obstacles, the tracking model is guaranteed to be safe.
Due to the random nature of RRT, during the simulation the system appears to randomly explore to look for an unobstructed path to the obstacle; we did not implement any exploration algorithms.

Fig. \ref{fig:simRRT_combined} shows three different time snapshots of the simulation.
At $t=8$, the planning model has sensed a portion of the previously unknown obstacles, and replans, so that the path deviates from a straight line from the initial position to the goal position.
The subplot showing $t = 47.7$ is rotated to show the trajectory up to this time from a more informative view angle.
Here, the system has safely passed by the first planar obstacle, and is moving around the second.
Note that the TEB never intersects the obstacles, implying that the tracking model is guaranteed to avoid collision with the obstacles, since it is guaranteed to stay within the TEB.
At $t=83.5$, the autonomous system safely passes by the last obstacle.

\begin{figure}
\centering
  \includegraphics[width=\columnwidth]{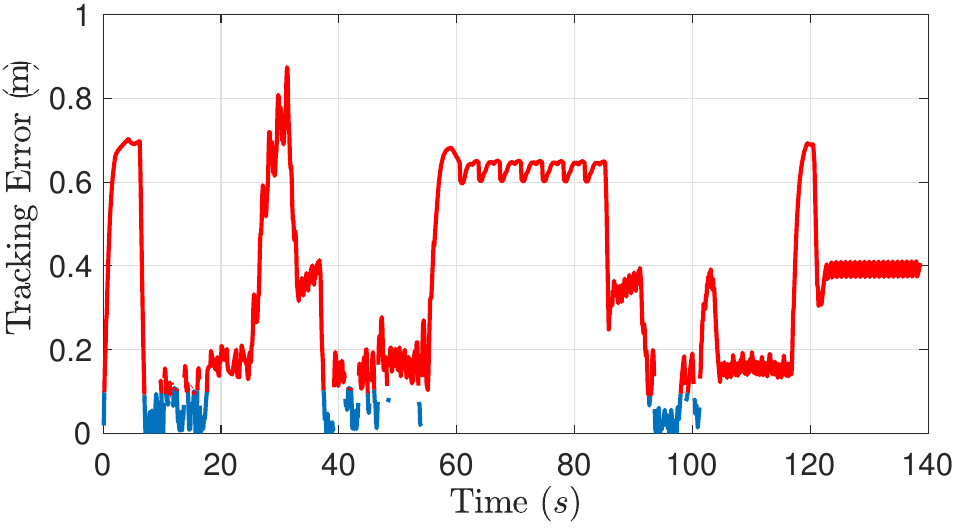}
  \caption{Tracking error over time for the 10D-3D example. The red dots indicate that the optimal tracking controller in \eqref{eq:opt_ctrl_inf} is used, while the blue dots indicate that an LQR controller for the linearized system is used. The tracking error stays below the predicted TEB of 0.9 m, despite worst-case wind.}
  \label{fig:tracking_error_RRT}
\end{figure}

Fig. \ref{fig:tracking_error_RRT} shows the maximum tracking error, in the three positional dimensions over time.
The red points indicate the time points at which the optimal tracking controller from Eq. \eqref{eq:opt_ctrl_inf} was used; this is the optimal tracking controller depicted in Fig. \ref{fig:hybrid_ctrl}.
The blue points indicate the time points at which a performance controller, also depicted in Fig. \ref{fig:hybrid_ctrl}, was used.
For the performance controller, we used a simple proportional controller that depends on the tracking error in each positional dimension; this controller is used whenever the tracking error is less than a quarter of the TEB.
From Fig. \ref{fig:tracking_error_RRT}, one can observe that the tracking error is always less than the TEB implied by the value function.

The simulation was done in MATLAB on a desktop computer with an Intel Core i7 2600K CPU.
The time was discretized in increments of 0.01.
On average per iteration, planning with RRT using a simple multi-tree RRT planning algorithm implemented in MATLAB modified from \cite{Gavin2013} took 5 ms, and computing the tracking controller took 5.5 ms.

\subsection{8D quadrotor-4D double integrator example with MPC \label{sec:resultsMPC}}

In this section, we demonstrate the online computation framework in Alg. \ref{alg:algOnline} with an 8D quadrotor example and MPC as the online planning algorithm.
Unlike in Sections \ref{sec:reach_planner} and \ref{sec:resultsRRT}, we consider a planning-tracking model pair for which the value funciton does not converge, so that the computation instead provides a tvTEB. In addition, the TEB depends on both position and speed, as opposed to just position. This is to accomodate velocity bounds on the system.

First we define the 8D dynamics of the near-hover quadrotor, and the 4D dynamics of a double integrator, which serves as the planning model to be used in MPC:

\begin{equation}
\label{eq:Quad8D_dyn}
\begin{bmatrix}
\dot x\\
\dot v_x\\
\dot \theta_x\\
\dot \omega_x\\
\dot y\\
\dot v_y\\
\dot \theta_y\\
\dot \omega_y
\end{bmatrix} =
\begin{bmatrix}
v_{x,s} + d_x\\
g \tan \theta_x\\
-d_1 \theta_x + \omega_x\\
-d_0 \theta_x + n_0 a_x\\
v_y + d_y\\
g \tan \theta_y\\
-d_1 \theta_y + \omega_y\\
-d_0 \theta_y + n_0 a_y
\end{bmatrix}, \quad
\begin{bmatrix}
\dot {\hat x}\\
\dot {\hat v}_x\\
\dot {\hat y}\\
\dot {\hat v}_y\\
\end{bmatrix} =
\begin{bmatrix}
\hat v_x\\
\hat a_x\\
\hat v_y\\
\hat a_y\\
\end{bmatrix},
\end{equation}

\noindent where the states, controls, and disturbances are the same as the first 8 components of the dynamics in \eqref{eq:Quad10D_dyn}.
The position $(\hat x,\hat y)$ and velocity $(\hat v_x, \hat v_y)$ are the states of the 4D system.
The controls are $(\hat a_x, \hat a_y)$, which represent the acceleration in each positional dimension.
The model parameters are chosen to be $d_0=10$, $d_1=8$, $n_0=10$, $k_T=0.91$, $g=9.81$, $|u_x|, |u_y| \le \pi/9$, $|\hat a_x|, |\hat a_y| \le 1$, $|\dstb_x|, |\dstb_y| \le 0.2$.

\subsubsection{Offline precomputation}
We define the relative system states to be the error states $(x_r, v_{x,r}, y_r, v_{y,r})$, which are the relative position and velocity, concatenated with the rest of the states in the 8D system.
Defining $\rtrans = \mathbf I_8$ and

\begin{equation*}
\small
\ptmat =
\begin{bmatrix}
  \begin{bmatrix} 1 \\ \mathbf 0_{3 \times 1} \end{bmatrix}
    & \mathbf 0_{4\times 1} \\
  \mathbf 0_{4\times 1}
    & \begin{bmatrix} 1 \\ \mathbf 0_{3 \times 1} \end{bmatrix}
\end{bmatrix},
\end{equation*}

\noindent we obtain the following relative system dynamics:

\begin{equation}
\label{eq:Quad8DRel_dyn}
\small
\begin{bmatrix}
\dot x_r\\
\dot v_{x,r}\\
\dot \theta_x\\
\dot \omega_x\\
\dot y_r\\
\dot v_{y,r}\\
\dot \theta_y\\
\dot \omega_y\\
\end{bmatrix} =
\begin{bmatrix}
v_{x,r} + \dstb_x\\
g \tan \theta_x - \hat a_x\\
-d_1 \theta_x + \omega_x\\
-d_0 \theta_x + n_0 a_x\\
v_{y,r} + \dstb_y\\
g \tan \theta_y - \hat a_y\\
-d_1 \theta_y + \omega_y\\
-d_0 \theta_y + n_0 a_y\\
\end{bmatrix}.
\end{equation}

As in the 10D-3D example in Section \ref{sec:resultsRRT}, the relative dynamics are decomposable into two 4D subsystems, and so computations were done in 4D space.

Fig. \ref{fig:vf_TEB:8D4D} shows the $(x_r, v_{x,r})$-projection of value function across several different times on the left subplot.
The total time horizon was $T=15$, and the value function did not converge.
The gray horizontal plane indicates the value of $\underline V$, which was $1.14$.
Note that with increasing $\tau$, $V(\rstate,\thor-\tau)$ is non-increasing, as proven in Prop. \ref{prop:nonconv}.
The right subplot of Fig. \ref{fig:vf_TEB:8D4D} shows the $(x_r, v_{x,r})$-projection of the tvTEB.
At $\tau=0$, the TEB is the smallest, and as $\tau$ increases, the size of TEB also increases, consistent with Proposition \ref{prop:nonconv}.
In other words, the set of possible error states $(x_r, v_{x,r})$ in the relative system increases with time, which makes intuitive sense.

The tvTEB shown in Fig. \ref{fig:vf_TEB:8D4D} is used to augment planning constraints in the $\hat x$ and $\hat v_x$ dimensions.
Since we have chosen identical parameters for the first four and last four states, the TEB in the $\hat y$ and $\hat v_y$ dimensions is identical.

On a desktop computer with an Intel Core i7 5820K CPU, the offline computation on a $81\times81\times65\times65$ grid with $75$ time points took approximately 30 hours and  required approximately 17 GB of RAM.
Note that unlike the other numerical examples, look-up tables representing the value function and its gradient must be stored at each time discretization point.

\subsubsection{Online sensing and planning}

\noindent We use the MPC design in \cite{Zhang2017} for online trajectory planning. It is formulated as a finite-time optimal control problem (OCP) given by

\begin{subequations} \label{eq:MPC}
  \begin{align}
  \underset{\{\bar{p}_k\}_{k=0}^N, \{\bar{u}_k\}_{k=0}^{N-1}}{\text{minimize}} \quad
  & \sum^{N-1}_{k=0} l(\bar{p}_k,\bar{u}_k) + l_f(\bar{p}_N-p_f)  \\
  \text{subject to} \quad & \bar{p}_0 = p_{\text{init}},\\
  &\bar{p}_{k+1} = h(\bar{p}_k,\bar{p}_k),\\
  & \label{eq:MPC:aug_constr} \bar{u}_k \in \mathbb{U},\enspace \bar{p}_k \in \constrAug(t_k),
  \end{align}
\end{subequations}
where $l(\cdot,\cdot)$ and $l_f(\cdot)$ are convex stage and terminal cost functions, the future states $\bar{p}_k$ and inputs $\bar{u}_k$ are decision variables, and $N$ is the prediction horizon. The index $t_k = t_0 + k \Delta t_{\text{mpc}}$ denotes for the internal time steps used in MPC, with $t_0$ and $\Delta t_{\text{mpc}}$ being the current time and the sampling interval of MPC, respectively.
Note that $N$ and $\Delta t_{\text{mpc}}$ are chosen such that $t_0 + N \Delta t_{\text{mpc}}\leq T - \tau$ with $\tau$ given by Step~9 in Algorithm~\ref{alg:algOnline} and $T$ defined for TEB in (\ref{eq:TEB_fin}).
States of the planning model are denoted by the variable $p = (\hat x, \hat v_x, \hat y, \hat v_y)$ with current state $p_{\text{init}}$ and terminal state $p_f$.
The planning control is denoted by the variable $u = (\hat a_x, \hat a_y)$.
The dynamical system $h(\cdot,\cdot)$ is set to be a zero-order hold discretized model of the 4D dynamics in \eqref{eq:Quad8D_dyn}. The state and input constraints are $\constrAug(t_k)$ and $\mathbb{U}$. Note that the time-varying constraint $\constrAug(t_k)$ contains the augmented state constraints:

\begin{equation}
\bar{p}_k \in \mathbb{P}_k :=\mathbb{P}\ominus\TEB_\estate(t_k),
\end{equation}

\noindent where $\mathbb{P}$ denotes the original state constraint, and $\TEB_\estate(t_k)$ is the tracking error bound at $t_k$.

For this example, we represent the augmented constraints as the complement of polytopes, which makes the MPC problem non-convex.
We follow the approach presented in \cite{Zhang2017} to compute a locally optimal solution that uses extra auxiliary variables for each non-convex constraint. We use a horizon $N=8$ with sampling interval $\Delta t_{\text{mpc}} = 0.2$ s. The MPC replans every 0.8 s. The control frequency is 10 Hz, i.e. $\Delta t = 0.1$ for both the 4D planning and 8D tracking system.

\begin{figure}
  \centering
  \begin{subfigure}[t]{0.49\columnwidth}
    \includegraphics[width=\columnwidth]{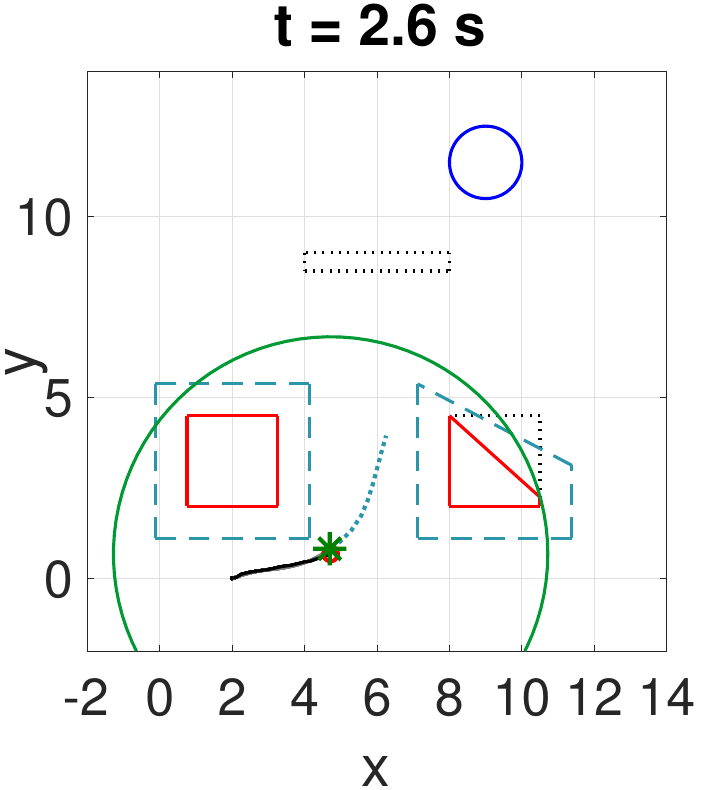}
  \end{subfigure}
  \begin{subfigure}[t]{0.49\columnwidth}
    \includegraphics[width=\columnwidth]{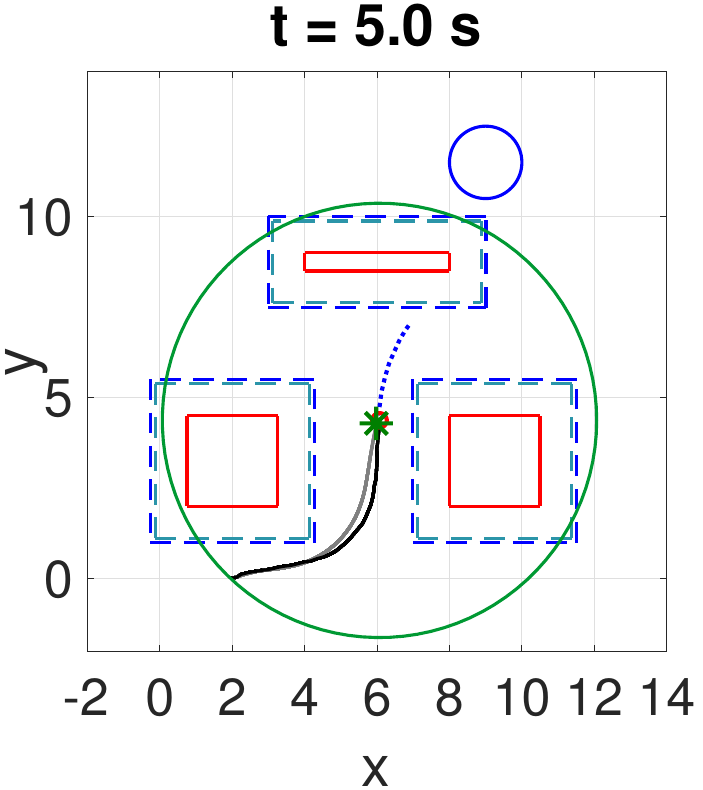}
  \end{subfigure}

  \begin{subfigure}[t]{0.49\columnwidth}
    \includegraphics[width=\columnwidth]{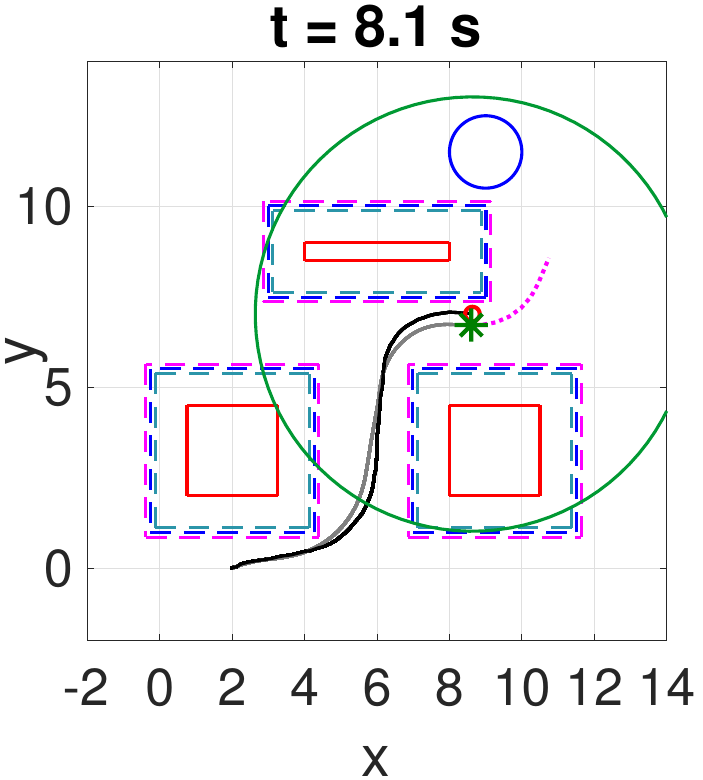}
  \end{subfigure}
  \begin{subfigure}[t]{0.49\columnwidth}
    \includegraphics[width=\columnwidth]{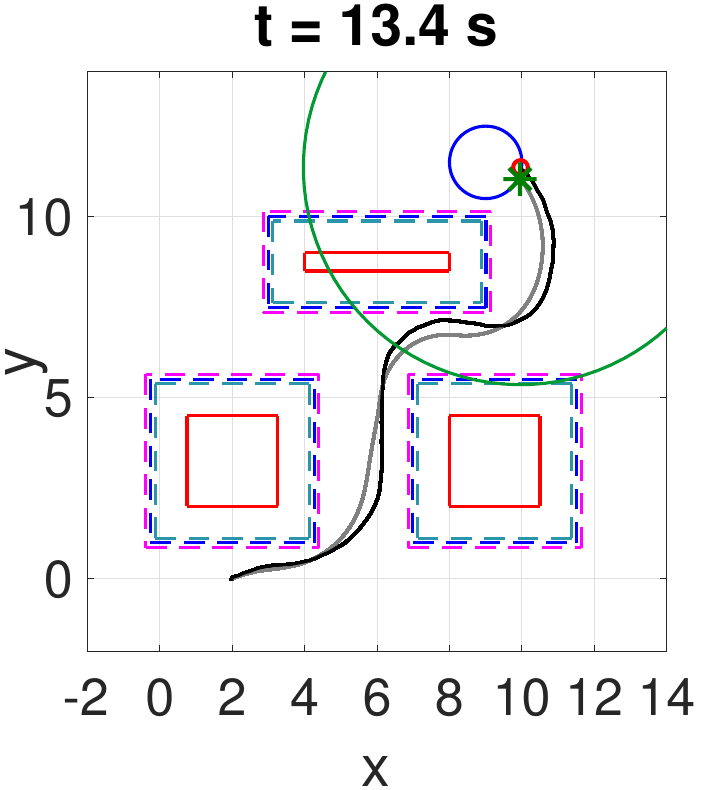}
  \end{subfigure}
  \caption{Simulation of the 8D-4D example. As the quadrotor with 8D dynamics (position shown as red circle and trajectory shown in black) senses new obstacles, the 4D planning system (position shown as green star and trajectory shown in grey) replans the trajectory, which is robustly tracked by the 8D system. The time-varying augmented obstacles are plotted as dashed polytopes with different sizes and colors.}

  \label{fig:8D4Dsim}
\end{figure}

The simulation showing the 8D quadrotor tracking the 4D double integrator is presented in Fig. \ref{fig:8D4Dsim}. The quadrotor starts at $(2.0,0.0)$ and seeks to reach the goal, i.e. the blue circle centered at $(9.0,11.5)$ with a radius of 1.0. Three initially unknown polytopic obstacles make up the constraints $\constr$.

The quadrotor has a circular sensing region with a radius of 6, colored in green. Unknown obstacles are marked with dotted black boundaries.
As the quadrotor explores the environment, obstacles are detected once they are within the sensing range.
The union of all sensed obstacles makes up the sensed constraints $\constrSense$, whose boundaries are colored in red.
They are enclosed by the augmented obstacles shown as dashed polytopes with different colors, each representing a different augmentation of the original sensed obstacles.
These illustrate the tvTEB in the position dimensions.
The tvTEB is also defined in the velocity dimensions; the value of the bounds over time are shown in Fig. \ref{fig:vf_TEB:8D4D}.
Therefore, time-varying constraints $\constrAug(t_k)$ in \eqref{eq:MPC:aug_constr} are augmented accordingly in all four dimensions of the planning system.

The MPC planner takes as input the current state $p_{\text{init}}$ of the 4D planning system and a list of augmented constraints $\constrAug(t_k)$. It then solves the OCP \eqref{eq:MPC} in real time for a sequence of optimized control inputs, which steers the planning system towards the goal while avoiding the augmented obstacles. The current state of the planning system is represented as a green star. In front of it, the predicted trajectories in $(\hat{x},\hat{y})$ space are plotted in dotted curves with the same color as the augmented obstacles considered at the time the MPC problem is solved. The traveled path of the planning system is shown as a solid grey curve. Here, unlike the standard MPC algorithm in which only the first element of the control inputs is used before replanning happens, our proposed MPC planner applies multiple control inputs to the 4D system in open-loop before it replans. This is due to the fact that the planning and tracking system use the same control frequency, i.e. they both update states every $\Delta t$ time steps as shown in Step~28 in Algorithm \ref{alg:algOnline}; and that the sampling interval of the planner $\Delta t_{\text{mpc}}$ is in general larger than $\Delta t$.

The state of the 8D tracking model \eqref{eq:Quad8DRel_dyn} is represented as a red circle. Using the hybrid tracking controller depicted in Fig. \ref{fig:hybrid_ctrl}, the 8D system tracks the 4D planning system within the tvTEB in Fig. \ref{fig:vf_TEB:8D4D}, which guarantees constraint satisfaction despite the tracking error at all times. The traveled path of the 8D system is shown as the solid black curve.

Fig. \ref{fig:8D4Dsim} also shows four time snapshots of the closed-loop simulation. The top left subplot shows the positional trajectories of the planning and tracking system at $t=2.6$. At $t=5.0$, the MPC planning algorithm speeds up and makes a sharp turn into a narrow corridor.
This results in a significant deviation between the two trajectories. However, the tracking controller keeps the system within the TEB and no collision is incurred at this time, as shown in the top right subplot.
A similar case can be observed in the lower left subplot.
Finally, at $t=13.4$ the quadrotor safely arrives at the destination.
Note that the size of the augmented obstacles keeps changing from time step to time step, as indicated by the dashed polytopes with different colors and sizes in the snapshots.
Meanwhile the MPC incorporates the updated information of augmented constraints $\constrAug(t_k)$ and plans safe trajectories.

\begin{figure}
  \includegraphics[width=0.95\columnwidth]{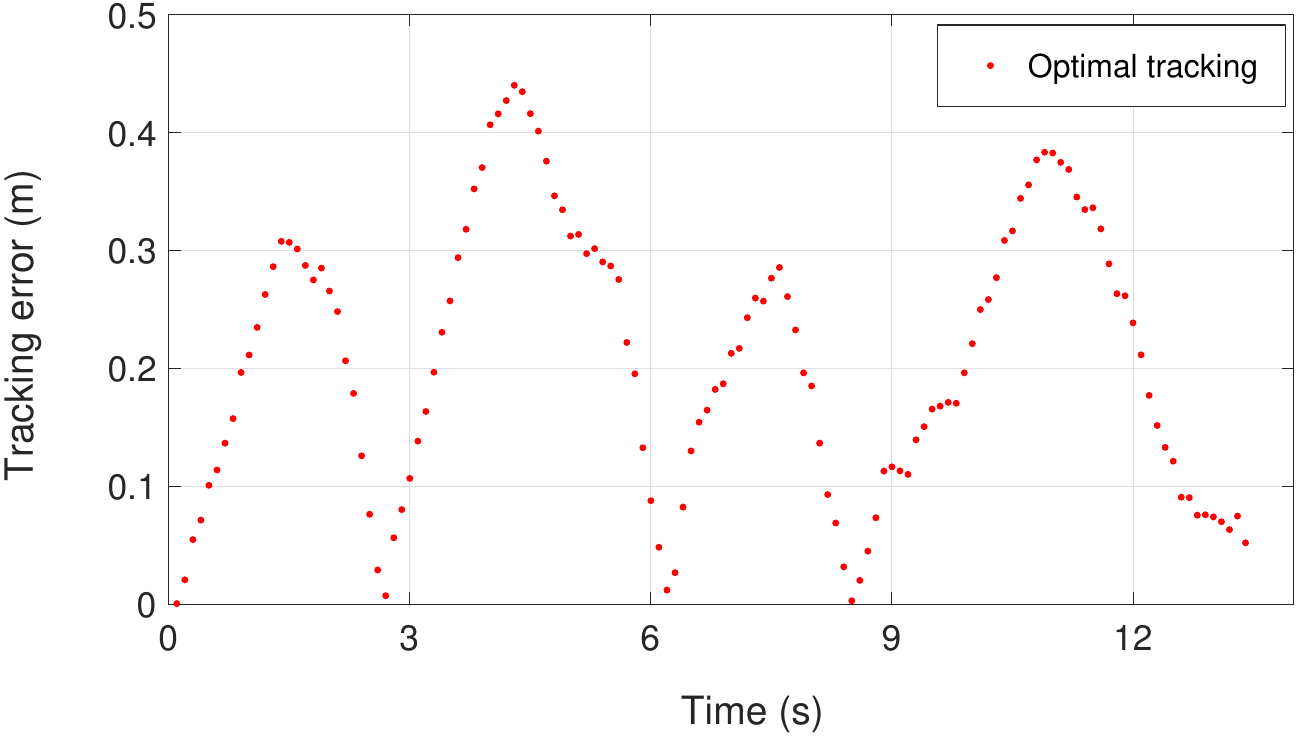}
  \centering
  \caption{Tracking error in $x_{r}$ over time for the 8D-4D example. The red dots represent the time steps when the optimal tracking controller in \eqref{eq:opt_ctrl_fin} is used. The tracking error remains lower than the minimal TEB of 1.11m at all times.}
  \label{fig:Q8D_tracking_error}
\end{figure}

Fig. \ref{fig:Q8D_tracking_error} shows the tracking error in $x_r$ over time. The red dots indicate the time points at which the optimal tracking controller from Eq. \eqref{eq:opt_ctrl_fin} is used. One may observe that the tracking error is always less than 0.45, well below the minimal TEB of 1.11 implied by the value function in Fig. \ref{fig:vf_TEB:8D4D}.

\begin{figure}
  \includegraphics[width=0.85\columnwidth]{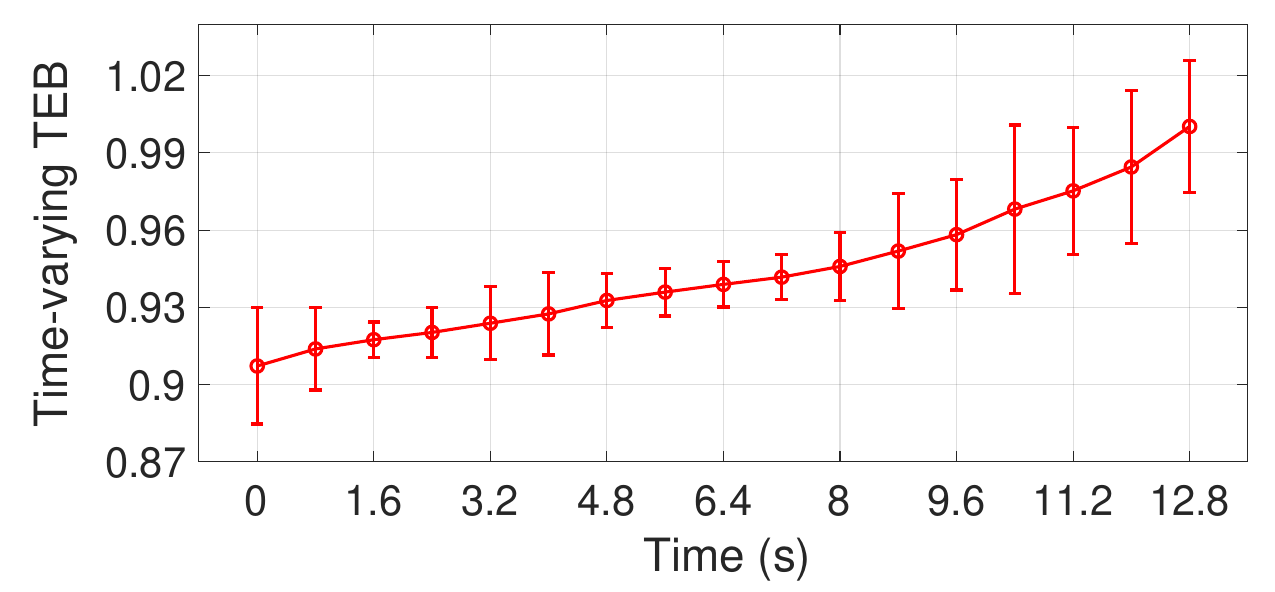}
  \centering
  \caption{Time-varying TEB over time in the $v_{x,r}$ dimension used by the MPC planner. The error bars are plotted at time steps when the MPC replans. They characterize the range of TEBs used for each planning while the red dots shows the mean value.}

  \label{fig:Q8D_TVTEB}
\end{figure}

Fig. \ref{fig:Q8D_TVTEB} shows the tvTEB in the $v_{x,r}$ dimension used in the closed-loop simulation.
Each time when replanning is performed via solving the MPC problem in Eq. \eqref{eq:MPC}, a sequence of 8 TEBs are determined by Step 9 in Algorithm \ref{alg:algOnline}, and used for constructing the augmented constraints in Eq. \eqref{eq:MPC:aug_constr}. Each error bar and the red dot in the middle shows the range and mean value of the TEBs used within a single MPC planning loop. Enabling a tvTEB allows us to set smaller TEBs for the initial time steps, growing the TEB as the time horizon extends.  If we were to use a fixed TEB, we would have to keep the error bound as the largest bound required over the time horizon; this can lead to fairly conservative movement from the planning algorithm. Note that the tvTEB is employed on full states including both position and velocity of the planning system, which allows us to reduce the conservativeness of the planner as much as possible.

We compare the simulation time where the tvTEB is used for planning with the case in that the TEB is fixed as a constant. The results are summarized in Table \ref{tab:sim_time}. One may observe that with the use of tvTEB the conservativeness in planning is reduced and a shorter flight time is achieved.

\begin{table}[htbp!]
\caption{\small{Simulation time with constant and tvTEB.}}
\centering
\normalsize
\begin{tabular}{ccccc}
\hline
&Simulation case &  Time (s)    \\
\hline
&Planning with constant TEB     &  14.6  \\
&Planning with tvTEB &  13.4  \\
\hline
\end{tabular}
\label{tab:sim_time}
\end{table}

Implementation of the MPC planner was based on MATLAB and \texttt{ACADO Toolkit} \cite{Houska2011a}. The nonlinear MPC problem was solved using an online active set strategy implemented in \texttt{qpOASES} \cite{Ferreau2014}. All the simulation results were obtained on a laptop with Ubuntu 14.04 LTS operating system and a Core i5-4210U CPU. The average computation time for solving the MPC problem \eqref{eq:MPC} was 0.37 s.
\section{Conclusions and Future work}
This paper introduced FaSTrack: Fast and Safe Tracking, a framework for providing trajectory-independent tracking error bounds (TEB) for a tracking model representing an autonomous system, and a planning model representing a simplified model of the autonomous system.
The TEB is obtained by analyzing a pursuit-evasion game between the the two models, and obtaining the associated value function by solving a Hamilton-Jacobi (HJ) variational inequality. When this game converges we acquire an infinite-horizon TEB, otherwise we acquire a time-varying TEB.

We demonstrated the framework's utility in three representative numerical simulations involving a 5D car model tracking a 3D car model planning using the fast sweeping method, a 10D quadrotor model tracking a 3D single integrator model planning using rapid-exploring random trees, and an 8D quadrotor model tracking a 4D double integrator planning using model predictive control. We considered simulated environments with static obstacles, but FaSTrack has been demonstrated in hardware to safely navigate around environments with static obstacles \cite{fridovich2018} and moving human pedestrians \cite{fisac2018probabilistically}.

There are still challenges and simplifications to address. The offline computation can be computationally prohibitive, a challenge we are working to address using techniques such as HJ reachability decomposition \cite{Chen2016DecouplingJournal,Chen2016DecouplingApprox}, sophisticated optimization techniques \cite{SinghChenEtAl2018}, and approximate dynamic programming \cite{royo2018classification}.  We are also interested in exploring methods to reduce conservativeness of the TEB by relaxing the worst-case assumption on the goals of the planning control.  Additionally, identifying when a particular tracking-planning model will have a converged value function remains an open question.  Finally, we currently assume both perfect sensing and a perfectly representative tracking model.  We would like to alleviate these assumptions by bounding uncertainty from sensing error, and making online updates to the value function as information about the tracking model is improved.

By computing trajectory-independent TEB, our framework decouples robustness guarantees from planning, and achieves the best of both worlds -- formal robustness guarantees which is usually computationally expensive to obtain, and real-time planning which usually sacrifices robustness.
Combined with any planning method in a modular fashion, our framework enables guaranteed safe planning and replanning in unknown environments, among other potential applications.

\bibliographystyle{IEEEtran}
\bibliography{references}

\begin{thebibliography}{10}
\providecommand{\url}[1]{#1}
\csname url@samestyle\endcsname
\providecommand{\newblock}{\relax}
\providecommand{\bibinfo}[2]{#2}
\providecommand{\BIBentrySTDinterwordspacing}{\spaceskip=0pt\relax}
\providecommand{\BIBentryALTinterwordstretchfactor}{4}
\providecommand{\BIBentryALTinterwordspacing}{\spaceskip=\fontdimen2\font plus
\BIBentryALTinterwordstretchfactor\fontdimen3\font minus
  \fontdimen4\font\relax}
\providecommand{\BIBforeignlanguage}[2]{{%
\expandafter\ifx\csname l@#1\endcsname\relax
\typeout{** WARNING: IEEEtran.bst: No hyphenation pattern has been}%
\typeout{** loaded for the language `#1'. Using the pattern for}%
\typeout{** the default language instead.}%
\else
\language=\csname l@#1\endcsname
\fi
#2}}
\providecommand{\BIBdecl}{\relax}
\BIBdecl

\bibitem{herbert2017fastrack}
S.~L. Herbert*, M.~Chen*, S.~Han, S.~Bansal, J.~F. Fisac, and C.~J. Tomlin,
  ``Fastrack: a modular framework for fast and guaranteed safe motion
  planning,'' \emph{Proc. IEEE Conf. Decision and Control}, 2017.

\bibitem{Mitchell05}
I.~Mitchell, A.~Bayen, and C.~Tomlin, ``A time-dependent {Hamilton-Jacobi}
  formulation of reachable sets for continuous dynamic games,'' \emph{IEEE
  Transactions on Automatic Control}, vol.~50, no.~7, pp. 947--957, July 2005.

\bibitem{SinghChenEtAl2018}
S.~Singh, M.~Chen, S.~L. Herbert, C.~J. Tomlin, and M.~Pavone, ``Robust
  tracking with model mismatch for fast and safe planning: an {SOS}
  optimization approach,'' in \emph{Workshop on Algorithmic Foundations of
  Robotics}, 2018.

\bibitem{royo2018classification}
V.~R. Royo, D.~Fridovich-Keil, S.~Herbert, and C.~J. Tomlin,
  ``Classification-based approximate reachability with guarantees applied to
  safe trajectory tracking,'' in \emph{Proc. IEEE Int. Conf. Robotics and
  Automation}, 2019.

\bibitem{Takei2013}
R.~Takei and R.~Tsai, ``{Optimal Trajectories of Curvature Constrained Motion
  in the Hamilton–Jacobi Formulation},'' \emph{J. Scientific Computing},
  vol.~54, no. 2-3, pp. 622--644, Feb. 2013.

\bibitem{Kuffner2000}
J.~J. Kuffner and S.~M. LaValle, ``Rrt-connect: An efficient approach to
  single-query path planning,'' in \emph{IEEE Int. Conf. Robotics and
  Automation}, 2000.

\bibitem{Kavraki1996}
L.~E. Kavraki, P.~Svestka, J.-C. Latombe, and M.~H. Overmars, ``Probabilistic
  roadmaps for path planning in high-dimensional configuration spaces,''
  \emph{IEEE Trans. Robotics and Automation}, vol.~12, no.~4, pp. 566--580,
  1996.

\bibitem{Qin2003}
S.~J. Qin and T.~A. Badgwell, ``A survey of industrial model predictive control
  technology,'' \emph{Control engineering practice}, vol.~11, no.~7, pp.
  733--764, 2003.

\bibitem{Zhang2017}
\BIBentryALTinterwordspacing
X.~Zhang, A.~Liniger, and F.~Borrelli, ``{Optimization-Based Collision
  Avoidance},'' Nov. 2017. [Online]. Available:
  \url{http://arxiv.org/abs/1711.03449}
\BIBentrySTDinterwordspacing

\bibitem{Hoy2015}
M.~Hoy, A.~S. Matveev, and A.~V. Savkin, ``Algorithms for collision-free
  navigation of mobile robots in complex cluttered environments: a survey,''
  \emph{Robotica}, vol.~33, no.~03, pp. 463--497, 2015.

\bibitem{Janson2015}
L.~Janson, E.~Schmerling, A.~Clark, and M.~Pavone, ``Fast marching tree: A fast
  marching sampling-based method for optimal motion planning in many
  dimensions,'' \emph{Int. J. Robotics Research}, vol.~34, no.~7, pp. 883--921,
  2015.

\bibitem{Richter2016}
C.~Richter, A.~Bry, and N.~Roy, ``Polynomial trajectory planning for aggressive
  quadrotor flight in dense indoor environments,'' in \emph{Robotics Research},
  2016.

\bibitem{Karaman2011}
S.~Karaman, M.~R. Walter, A.~Perez, E.~Frazzoli, and S.~Teller, ``Anytime
  motion planning using the rrt,'' in \emph{IEEE Int. Conf. Robotics and
  Automation}, 2011.

\bibitem{Kobilarov2012}
M.~Kobilarov, ``Cross-entropy motion planning,'' \emph{Int. J. Robotics
  Research}, vol.~31, no.~7, pp. 855--871, 2012.

\bibitem{Schulman2013}
J.~Schulman, J.~Ho, A.~X. Lee, I.~Awwal, H.~Bradlow, and P.~Abbeel, ``Finding
  locally optimal, collision-free trajectories with sequential convex
  optimization.'' in \emph{Proc. Robotics: science and systems}, 2013.

\bibitem{Ratliff2009}
N.~Ratliff, M.~Zucker, J.~A. Bagnell, and S.~Srinivasa, ``Chomp: Gradient
  optimization techniques for efficient motion planning,'' in \emph{IEEE Int.
  Conf. Robotics and Automation}, 2009.

\bibitem{Vitus2008}
M.~Vitus, V.~Pradeep, G.~Hoffmann, S.~Waslander, and C.~Tomlin, ``Tunnel-milp:
  Path planning with sequential convex polytopes,'' in \emph{AIAA Guidance,
  Navigation and Control Conf. and Exhibit}, 2008.

\bibitem{Zeilinger2011}
M.~N. Zeilinger, C.~N. Jones, and M.~Morari, ``Real-time suboptimal model
  predictive control using a combination of explicit mpc and online
  optimization,'' \emph{IEEE Trans. Autom. Control}, vol.~56, no.~7, pp.
  1524--1534, 2011.

\bibitem{Richter2012}
S.~Richter, C.~N. Jones, and M.~Morari, ``Computational complexity
  certification for real-time mpc with input constraints based on the fast
  gradient method,'' \emph{IEEE Trans. Autom. Control}, vol.~57, no.~6, pp.
  1391--1403, 2012.

\bibitem{Richards2006}
A.~Richards and J.~P. How, ``Robust variable horizon model predictive control
  for vehicle maneuvering,'' \emph{Int. J. Robust and Nonlinear Control},
  vol.~16, no.~7, pp. 333--351, 2006.

\bibitem{DiCairano2016}
S.~Di~Cairano and F.~Borrelli, ``Reference tracking with guaranteed error bound
  for constrained linear systems,'' \emph{IEEE Trans. Autom. Control}, vol.~61,
  no.~8, pp. 2245--2250, 2016.

\bibitem{Diehl2002}
M.~Diehl, H.~G. Bock, J.~P. Schl{\"o}der, R.~Findeisen, Z.~Nagy, and
  F.~Allg{\"o}wer, ``Real-time optimization and nonlinear model predictive
  control of processes governed by differential-algebraic equations,'' \emph{J.
  Process Control}, vol.~12, no.~4, pp. 577--585, 2002.

\bibitem{Schildbach2016}
G.~Schildbach and F.~Borrelli, ``A dynamic programming approach for
  nonholonomic vehicle maneuvering in tight environments,'' in \emph{IEEE
  Intelligent Vehicles Symposium}, 2016.

\bibitem{Diehl2009}
M.~Diehl, H.~J. Ferreau, and N.~Haverbeke, ``Efficient numerical methods for
  nonlinear mpc and moving horizon estimation,'' in \emph{Nonlinear Model
  Predictive Control}, 2009.

\bibitem{Neunert2016}
M.~Neunert, C.~de~Crousaz, F.~Furrer, M.~Kamel, F.~Farshidian, R.~Siegwart, and
  J.~Buchli, ``Fast nonlinear model predictive control for unified trajectory
  optimization and tracking,'' in \emph{IEEE Int. Conf. Robotics and
  Automation}, 2016.

\bibitem{Gillula2010}
J.~H. Gillula, H.~Huang, M.~P. Vitus, and C.~J. Tomlin, ``Design of guaranteed
  safe maneuvers using reachable sets: Autonomous quadrotor aerobatics in
  theory and practice,'' in \emph{IEEE Int. Conf. on Robotics and Automation},
  2010, pp. 1649--1654.

\bibitem{Dey2016}
D.~Dey, K.~S. Shankar, S.~Zeng, R.~Mehta, M.~T. Agcayazi, C.~Eriksen,
  S.~Daftry, M.~Hebert, and J.~A. Bagnell, ``Vision and learning for
  deliberative monocular cluttered flight,'' in \emph{Field and Service
  Robotics}.\hskip 1em plus 0.5em minus 0.4em\relax Springer, 2016, pp.
  391--409.

\bibitem{Majumdar2017}
A.~Majumdar and R.~Tedrake, ``{Funnel libraries for real-time robust feedback
  motion planning},'' \emph{Int. J. Robotics Research}, vol.~36, no.~8, pp.
  947--982, Jul 2017.

\bibitem{Kalakrishnan2011}
M.~Kalakrishnan, S.~Chitta, E.~Theodorou, P.~Pastor, and S.~Schaal, ``Stomp:
  Stochastic trajectory optimization for motion planning,'' in \emph{IEEE
  International Conference on Robotics and Automation}.\hskip 1em plus 0.5em
  minus 0.4em\relax IEEE, 2011, pp. 4569--4574.

\bibitem{Schwesinger2013}
U.~Schwesinger, M.~Rufli, P.~Furgale, and R.~Siegwart, ``A sampling-based
  partial motion planning framework for system-compliant navigation along a
  reference path,'' in \emph{Intelligent Vehicles Symposium (IV), 2013
  IEEE}.\hskip 1em plus 0.5em minus 0.4em\relax IEEE, 2013, pp. 391--396.

\bibitem{KousikVaskovEtAl2017}
S.~Kousik, S.~Vaskov, M.~Johnson-Roberson, and R.~Vasudevan, ``Safe trajectory
  synthesis for autonomous driving in unforeseen environments,'' in \emph{Proc.
  ASME Dynamic Systems and Control Conf.}, 2017.

\bibitem{Bansal2017}
S.~Bansal, M.~Chen, J.~F. Fisac, and C.~J. Tomlin, ``{Safe Sequential Path
  Planning of Multi-Vehicle Systems Under Presence of Disturbances and
  Imperfect Information},'' in \emph{Proc. American Control Conf.}, 2017.

\bibitem{Burridge1999}
R.~R. Burridge, A.~A. Rizzi, and D.~E. Koditschek, ``{Sequential Composition of
  Dynamically Dexterous Robot Behaviors},'' \emph{Int. J. Robotics Research},
  vol.~18, no.~6, pp. 534--555, Jun 1999.

\bibitem{Parrilo2000}
P.~A. Parrilo, ``{Structured semidefinite programs and semialgebraic geometry
  methods in robustness and optimization},'' Ph.D. Dissertation, California
  Institute of Technology, 2000.

\bibitem{Singh2017}
S.~Singh, A.~Majumdar, J.-J. Slotine, and M.~Pavone, ``Robust online motion
  planning via contraction theory and convex optimization,'' \emph{ICRA
  submission}, 2017.

\bibitem{Ahmadi2017}
A.~A. Ahmadi and A.~Majumdar, ``{DSOS and SDSOS Optimization: More Tractable
  Alternatives to Sum of Squares and Semidefinite Optimization},'' Jun 2017.

\bibitem{Ames2014}
A.~D. Ames, J.~W. Grizzle, and P.~Tabuada, ``Control barrier function based
  quadratic programs with application to adaptive cruise control,'' in
  \emph{IEEE Conference on Decision and Control}.\hskip 1em plus 0.5em minus
  0.4em\relax IEEE, 2014, pp. 6271--6278.

\bibitem{Xu2015}
X.~Xu, P.~Tabuada, J.~W. Grizzle, and A.~D. Ames, ``Robustness of control
  barrier functions for safety critical control** this work is partially
  supported by the national science foundation grants 1239055, 1239037 and
  1239085.'' \emph{IFAC-PapersOnLine}, vol.~48, no.~27, pp. 54--61, 2015.

\bibitem{Coddington84}
E.~A. Coddington and N.~Levinson, \emph{Theory of Ordinary Differential
  Equations}.\hskip 1em plus 0.5em minus 0.4em\relax Krieger Pub Co, 1984.

\bibitem{Fisac15}
J.~F. Fisac, M.~Chen, C.~J. Tomlin, and S.~S. Shankar, ``Reach-avoid problems
  with time-varying dynamics, targets and constraints,'' in \emph{18th
  International Conference on Hybrid Systems: Computation and Controls}, 2015.

\bibitem{Tomlin00}
C.~J. Tomlin, J.~Lygeros, and S.~S. Sastry, ``A game theoretic approach to
  controller design for hybrid systems,'' \emph{Proceedings of the IEEE},
  vol.~88, no.~7, pp. 949 --970, July 2000.

\bibitem{Chen2016DecouplingJournal}
M.~Chen, S.~L. Herbert, M.~Vashishtha, S.~Bansal, and C.~J. Tomlin,
  ``{Decomposition of Reachable Sets and Tubes for a Class of Nonlinear
  Systems},'' \emph{IEEE Trans. Autom. Control}, Nov. 2018.

\bibitem{Chen2018}
M.~Chen and C.~J. Tomlin, ``{Hamilton-Jacobi Reachability: Some Recent
  Theoretical Advances and Applications in Unmanned Airspace Management},''
  \emph{Annual Review of Control, Robotics, and Autonomous Systems}, vol.~1,
  2018.

\bibitem{Mitchell07c}
I.~M. Mitchell, ``{The Flexible, Extensible and Efficient Toolbox of Level Set
  Methods},'' \emph{J. Scientific Computing}, vol.~35, no. 2-3, pp. 300--329,
  Jun. 2008.

\bibitem{Mitchell2012}
I.~M. Mitchell, M.~Chen, and M.~Oishi, ``{Ensuring safety of nonlinear sampled
  data systems through reachability},'' \emph{IFAC Proc. Volumes}, vol.~45,
  no.~9, pp. 108--114, 2012.

\bibitem{Mitchell13}
I.~M. Mitchell, S.~Kaynama, M.~Chen, and M.~Oishi, ``{Safety preserving control
  synthesis for sampled data systems},'' \emph{Nonlinear Analysis: Hybrid
  Systems}, vol.~10, no.~1, pp. 63--82, Nov. 2013.

\bibitem{Dabadie2014}
C.~Dabadie, S.~Kaynama, and C.~J. Tomlin, ``{A practical reachability-based
  collision avoidance algorithm for sampled-data systems: Application to ground
  robots},'' in \emph{Proc. IEEE Int. Conf. on Intelligent Robots and Systems},
  Sept. 2014, pp. 4161--4168.

\bibitem{Akametalu2014}
A.~K. Akametalu, J.~F. Fisac, J.~H. Gillula, S.~Kaynama, M.~N. Zeilinger, and
  C.~J. Tomlin, ``{Reachability-based safe learning with Gaussian processes},''
  in \emph{Proc. IEEE Conf. on Decision and Control}, Dec. 2014.

\bibitem{bajcsy2019efficient}
A.~Bajcsy, S.~Bansal, E.~Bronstein, V.~Tolani, and C.~J. Tomlin, ``{An
  Efficient Reachability-Based Framework for Provably Safe Autonomous
  Navigation in Unknown Environments},'' in \emph{Proc. IEEE Conf. Decision and
  Control}, 2019.

\bibitem{fridovich2018safe}
D.~Fridovich-Keil, J.~F. Fisac, and C.~J. Tomlin, ``{Safely Probabilistically
  Complete Real-Time Planning and Exploration in Unknown Environments},'' in
  \emph{Proc. Int. Conf. Robotics and Automation}, 2019.

\bibitem{fisac2018probabilistically}
J.~F. Fisac, A.~Bajcsy, S.~L. Herbert, D.~Fridovich-Keil, S.~Wang, C.~J.
  Tomlin, and A.~D. Dragan, ``Probabilistically safe robot planning with
  confidence-based human predictions,'' in \emph{Proc. Robotics: Science and
  Systems}, 2018.

\bibitem{fridovich2019confidence}
D.~Fridovich-Keil, A.~Bajcsy, J.~F. Fisac, S.~L. Herbert, S.~Wang, A.~D.
  Dragan, and C.~J. Tomlin, ``Confidence-aware motion prediction for real-time
  collision avoidance,'' \emph{Int. J. Robotics Research}, 2019.

\bibitem{chen2018robust}
M.~Chen, S.~Bansal, J.~F. Fisac, and C.~J. Tomlin, ``Robust sequential
  trajectory planning under disturbances and adversarial intruder,'' \emph{IEEE
  Trans. Control Systems Technology}, vol.~27, no.~4, pp. 1566--1582, 2018.

\bibitem{bajcsy2018scalable}
A.~Bajcsy*, S.~L. Herbert*, D.~Fridovich-Keil, J.~F. Fisac, S.~Deglurkar, A.~D.
  Dragan, and C.~J. Tomlin, ``A scalable framework for real-time multi-robot,
  multi-human collision avoidance,'' \emph{IEEE Conf. Robotics and Automation},
  2018.

\bibitem{Bouffard12}
P.~Bouffard, ``On-board model predictive control of a quadrotor helicopter:
  Design, implementation, and experiments,'' Master's thesis, University of
  California, Berkeley, 2012.

\bibitem{Gavin2013}
\BIBentryALTinterwordspacing
{Gavin (Matlab community Contributor)}, ``{Multiple Rapidly-exploring Random
  Tree (RRT)},'' 2013. [Online]. Available:
  \url{https://www.mathworks.com/matlabcentral/fileexchange/21443-multiple-rapidly-exploring-random-tree--rrt-}
\BIBentrySTDinterwordspacing

\bibitem{Houska2011a}
B.~Houska, H.~Ferreau, and M.~Diehl, ``{ACADO} {T}oolkit -- {A}n {O}pen
  {S}ource {F}ramework for {A}utomatic {C}ontrol and {D}ynamic
  {O}ptimization,'' \emph{Optimal Control Applications and Methods}, vol.~32,
  no.~3, pp. 298--312, 2011.

\bibitem{Ferreau2014}
H.~Ferreau, C.~Kirches, A.~Potschka, H.~Bock, and M.~Diehl, ``qp{OASES}: A
  parametric active-set algorithm for quadratic programming,''
  \emph{Mathematical Programming Computation}, vol.~6, no.~4, pp. 327–--363,
  2014.

\bibitem{fridovich2018}
D.~Fridovich-Keil*, S.~L. Herbert*, J.~F. Fisac*, S.~Deglurkar, and C.~J.
  Tomlin, ``Planning, fast and slow: A framework for adaptive real-time safe
  trajectory planning.'' \emph{IEEE Conf. Robotics and Automation}, 2018.

\bibitem{Chen2016DecouplingApprox}
M.~Chen, S.~', and C.~J. Tomlin, ``Fast reachable set approximations via state
  decoupling disturbances,'' in \emph{IEEE Conference on Decision and
  Control}.\hskip 1em plus 0.5em minus 0.4em\relax IEEE, 2016, pp. 191--196.

\end{thebibliography}

\section*{Appendix: Infinite Time Horizon TEB}

When the value function \eqref{eq:valfunc} converges, we write $\valfunc(\thor, \rstate) := \valfunc_\infty(\rstate)$.
The optimal controller is then given by

\begin{align} \label{eq:opt_ctrl_inf}
\tctrl^*(\rstate) = \arg\min_{\tctrl\in\tcset} \max_{\pctrl\in\pcset, \dstb\in\dset} \nabla\valfunc(\rstate) \cdot \rdyn(\rstate,\tctrl,\pctrl,\dstb).
\end{align}

\noindent with the optimal planning control and disturbance given by

\begin{align} \label{eq:opt_dstb_inf}
\begin{bmatrix}
\pctrl^* \\
\dstb^*
\end{bmatrix}(\rstate) = \arg \max_{\pctrl\in\pcset, \dstb\in\dset} \nabla\valfunc_\infty(\rstate) \cdot \rdyn(\rstate,\tctrl^*,\pctrl,\dstb).
\end{align}

The smallest level set corresponding to the value $\underline\valfunc_\infty := \min_{\rstate} \valfunc_\infty(\rstate)$ can be interpreted as the smallest possible tracking error of the system, and
the TEB is given by the set
\begin{align} \label{eq:TEB_inf}
\TEB_\infty = \{\rstate: \valfunc_\infty(\rstate) \le \underline\valfunc_\infty\}.
\end{align}

\noindent with the TEB in error state subspace is given by
  \begin{align}  \label{eq:TEBp_inf}
  \TEB_{\estate, \infty} = \{\estate: \exists \astate, \valfunc_\infty(\estate, \astate) \le \underline\valfunc_\infty\}.
  \end{align}

In the online implementation in Alg. \ref{alg:algOnline}, one replaces all mentions of value function and TEB with their infinite time horizon counterpart, and skip Line \ref{ln:infSkip}.
Finally, Prop. \ref{prop:main} provides the infinite time horizon result analogous to Prop. \ref{prop:nonconv}.

\begin{prop}
  \label{prop:main}
  \textbf{Infinite time horizon guaranteed TEB}. Given $\tvar \ge 0$, $\forall \tvar' \ge \tvar, ~\rstate\in\TEB_\infty \Rightarrow \rtraj^*(\tvar'; \rstate, \tvar) \in \TEB_\infty$,  with $\rtraj^*$ defined the same way as in \eqref{eq:fin_thor_prop:here} to \eqref{eq:fin_thor_prop:there}.

\end{prop}

\textit{Proof:}
    Suppose that the value function converges, and define
  \begin{equation}
  \label{eq:conv_valfunc}
  \valfunc_\infty(\rstate) := \lim_{\thor\rightarrow\infty}\valfunc(\rstate, T)
  \end{equation}

  We first show that for all $\tvar, \tvar'$ with $\tvar' \ge \tvar$,
  \begin{equation}
  \label{eq:invariant}
  \valfunc_\infty(\rstate) \ge \valfunc_\infty(\rtraj^*(\tvar'; \rstate, \tvar)).
  \end{equation}

  Without loss of generality, assume $\tvar=0$. Then, we have

  \begin{subequations} \label{eq:inf_thor_steps}
    \begin{align}
    \valfunc_\infty(\rstate) & = \lim_{\thor\rightarrow\infty}\max_{\tau \in [0, \thor]} \errfunc(\rtraj^*(\tau; \rstate, 0)) \label{eq:inf_thor_steps:1}\\
    &= \lim_{\thor\rightarrow\infty}\max_{\tau \in [-\tvar', \thor]} \errfunc(\rtraj^*(\tau; \rstate, -\tvar')) \label{eq:inf_thor_steps:2}\\
    &\ge \lim_{\thor\rightarrow\infty}\max_{\tau \in [0, \thor]} \errfunc(\rtraj^*(\tau; \rstate, -\tvar')) \label{eq:inf_thor_steps:3}\\
    & = \lim_{\thor\rightarrow\infty}\max_{\tau \in [0, \thor]} \errfunc(\rtraj^*(\tau; \rtraj^*(0; \rstate, -\tvar'), 0)) \label{eq:inf_thor_steps:4}\\
    & = \lim_{\thor\rightarrow\infty}\max_{\tau \in [0, \thor]} \errfunc(\rtraj^*(\tau; \rtraj^*(\tvar'; \rstate, 0), 0)) \label{eq:inf_thor_steps:5}\\
    & = \valfunc_\infty(\rtraj^*(\tvar'; \rstate, 0)) \label{eq:inf_thor_steps:6}
    \end{align}
  \end{subequations}

  Explanation of steps:
  \begin{itemize}
    \item \eqref{eq:inf_thor_steps:1} and \eqref{eq:fin_thor_steps:6}: by definition of value function
    \item \eqref{eq:inf_thor_steps:2}: shifting time by $-\tvar'$
    \item \eqref{eq:inf_thor_steps:3}: removing the time interval $[-\tvar',0)$ in the $\max$ operator
    \item \eqref{eq:inf_thor_steps:4}: splitting trajectory $\rtraj^*(\tau; \rstate, -\tvar')$ into two stages corresponding to time intervals $[-\tvar', 0]$ and $[0, \tau]$
    \item \eqref{eq:inf_thor_steps:5}: shifting time reference in $\rtraj^*(0; \rstate, -\tvar')$ by $\tvar'$, since dynamics are time-invariant
  \end{itemize}

  Now, we finish the proof as follows:
  \begin{subequations} \label{eq:inf_hor}
    \begin{align}
    \rstate \in \TEB_\infty &\Leftrightarrow \valfunc_\infty(\rstate) \le \underline\valfunc \\
    & \Rightarrow \valfunc_\infty(\rtraj^*(\tvar'; \rstate, \tvar)) \le \underline\valfunc \label{eq:inf_hor:2}\\
    & \Leftrightarrow \rtraj^*(\tvar'; \rstate, \tvar) \in \TEB_\infty,
    \end{align}
  \end{subequations}

  \noindent where $\eqref{eq:invariant}$ is used for the step in \eqref{eq:inf_hor:2}. \hfill $\blacksquare$
\begin{IEEEbiography}[{\includegraphics[width=1in,height=1.25in,clip,keepaspectratio]{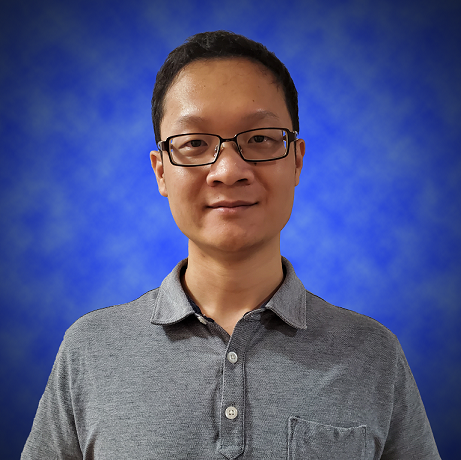}}]{Mo Chen}
    is an Assistant Professor in the School of Computing Science at Simon Fraser University, Burnaby, BC, Canada, where he directs the Multi-Agent Robotic Systems Lab. He completed his PhD in the Electrical Engineering and Computer Sciences Department at the University of California, Berkeley with Claire Tomlin in 2017, and received his BASc in Engineering Physics from the University of British Columbia in 2011. From 2017 to 2018, Mo was a postdoctoral researcher in the Aeronautics and Astronautics Department in Stanford University with Marco Pavone. His research interests include multi-agent systems, safety-critical systems, and the intersection between control theory and robotic learning.
\end{IEEEbiography}

\begin{IEEEbiography}[{\includegraphics[width=1in,height=1.25in,clip,keepaspectratio]{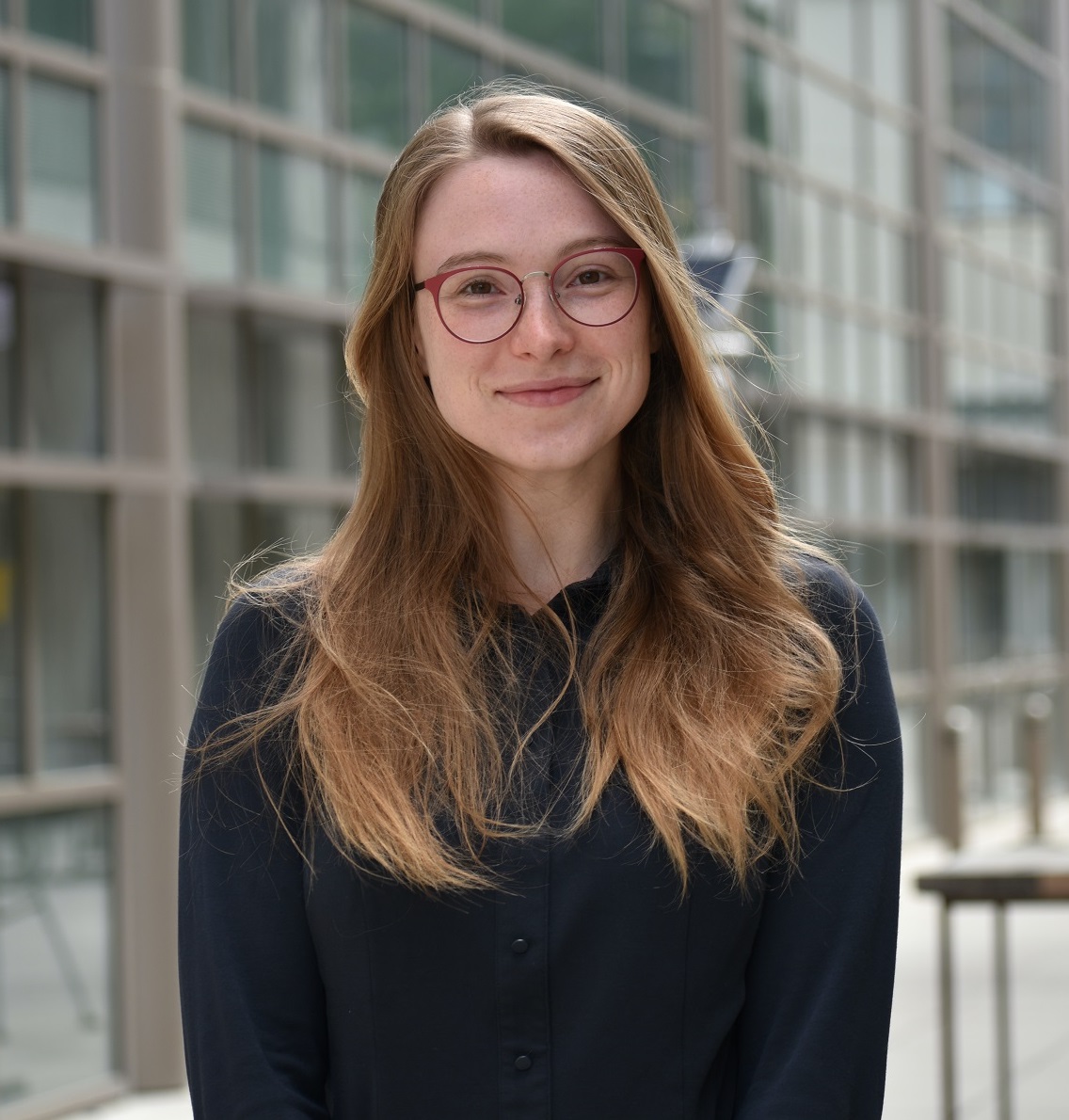}}]{Sylvia Herbert}
	is an Assistant Professor at UC San Diego, where she runs the Safe Autonomous Systems Lab. Her research is on efficiently guaranteeing safe control of autonomous systems based on available models and given information about the environment.  The lab uses tools from optimal control theory and dynamics games, machine learning, and cognitive science to develop new safety techniques that are able to quickly adapt to unexpected changes and new information in the system or the environment. Professor Herbert received her PhD from UC Berkeley Electrical Engineering and Computer Sciences in 2020, and BS/MS from Drexel University Mechanical Engineering in 2014.
\end{IEEEbiography}
\begin{IEEEbiography}[{\includegraphics[width=1in,height=1.25in,clip,keepaspectratio]{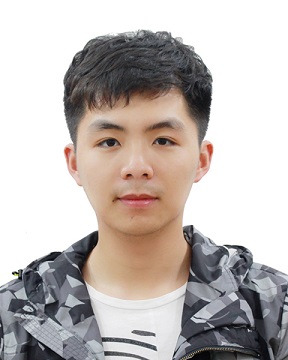}}]{Haimin Hu}
    is currently a Ph.D. student in Electrical Engineering at Princeton University. He received his B.E. degree in Electronic and Information Engineering from ShanghaiTech University in 2018, and an M.S.E. degree in Electrical Engineering from the University of Pennsylvania in 2020. From 2017 to 2018 he was a visiting undergraduate student in the Department of Electrical Engineering and Computer Sciences at the University of California, Berkeley. His research interests include learning for control, human-robot interaction, model predictive control, and multi-agent systems.
\end{IEEEbiography}
\addtolength{\textheight}{-8cm}   
\begin{IEEEbiography}[{\includegraphics[width=1in,height=1.25in,clip,keepaspectratio]{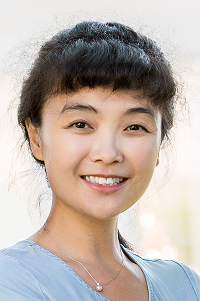}}]{Ye Pu}
	received the B.S. degree in electrical engineering from Shanghai Jiao Tong University, China, in 2008, the M.S. degree in electrical engineering and computer sciences from the Technical University Berlin, Germany, in 2011, and the PhD degree in electrical engineering from Swiss Federal Institute of Technology Lausanne (EPFL), Switzerland, in 2016. She was a Swiss NSF Early Postdoc.Mobility fellow and a postdoctoral researcher with the Department of Electrical Engineering and Computer Sciences at the University of California at Berkeley, CA, USA, from 2016 to 2018. She is a Lecturer (Assistant Professor) at the Department of Electrical and Electronic Engineering at the University of Melbourne, Australia. Her current research interests include learning-based control, optimisation algorithms and multi-agent systems with applications to underwater robotics and energy distribution systems.
\end{IEEEbiography}
\begin{IEEEbiography}[{\includegraphics[width=1in,height=1.25in,clip,keepaspectratio]{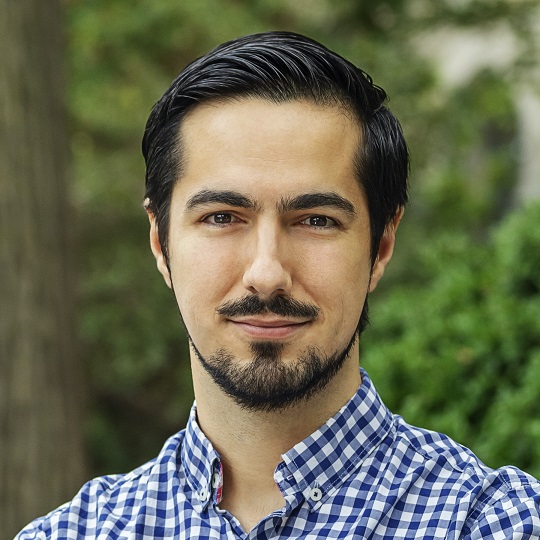}}]{Jaime Fern\'andez Fisac}
	 is an Assistant Professor of Electrical Engineering at Princeton University, where he directs the Safe Robotics Laboratory. His research interests lie in control theory and artificial intelligence, with a focus on safety for human-centered autonomy. From 2019 to 2020, he was a Research Scientist at Waymo, working on autonomous vehicle safety and interaction. He received a B.S./M.S. degree in Electrical Engineering from the Universidad Polit\'ecnica de Madrid, Spain, in 2012, a M.Sc. in Aeronautics from Cranfield University, U.K., in 2013, and his Ph.D. in Electrical Engineering and Computer Sciences at the University of California, Berkeley in 2019. He is a recipient of the ``la Caixa'' Foundation Fellowship and the Leon O. Chua award for outstanding achievement in nonlinear science.
\end{IEEEbiography}
\begin{IEEEbiography}[{\includegraphics[width=1in,height=1.25in,clip,keepaspectratio]{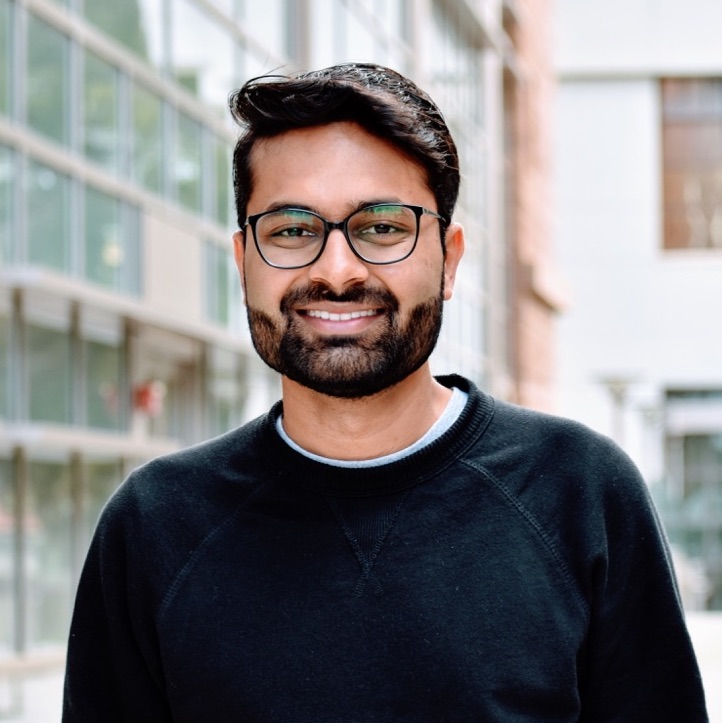}}]{Somil Bansal}
	is an Assistant Professor in the Department of Electrical and Computer Engineering at the University of Southern California, Los Angeles. He completed his MS and PhD in the Electrical Engineering and Computer Sciences Department at the University of California, Berkeley in 2014 and 2020 respectively, and received his B.Tech. in Electrical Engineering from Indian Institute of Technology, Kanpur in 2012. From 2020 to 2021, Somil was a research scientist at Waymo. His research interests include developing mathematical tools and algorithms for control and analysis of autonomous systems, with a focus on bridging learning and control-theoretic approaches for safety-critical autonomous systems.
\end{IEEEbiography}
\begin{IEEEbiography}[{\includegraphics[width=1in,height=1.25in,clip,keepaspectratio]{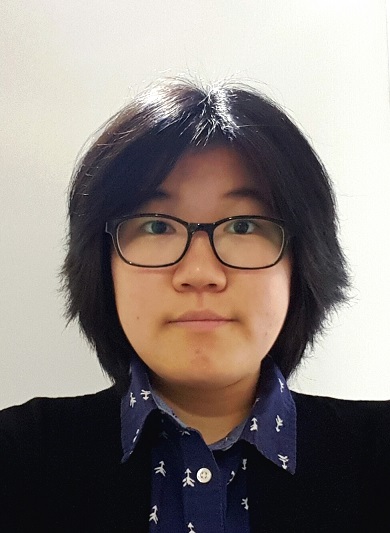}}]{SooJean Han}
	received her B.S. degree in Electrical Engineering and Computer Science, and Applied Mathematics at the University of California, Berkeley (UC Berkeley) in 2016. In 2017, she was a research assistant in the Hybrid Systems Lab at UC Berkeley, and a research assistant in the CAST Lab at California Institute of Technology (Caltech). She is currently a Ph.D. candidate in Control and Dynamical Systems at Caltech. Her research interests lie primarily in stochastic control and artificial intelligence. She is a recipient of the Caltech Special EAS Fellowship and the NSF GRFP.
\end{IEEEbiography}
\begin{IEEEbiography}[{\includegraphics[width=1in,height=1.25in,clip,keepaspectratio]{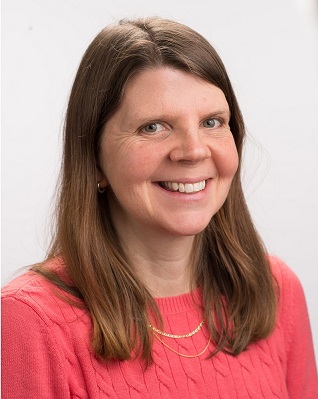}}]{Claire J. Tomlin}
	is the Charles A. Desoer Professor of Engineering in Electrical Engineering and Computer Sciences at the University of California, Berkeley. She was an Assistant, Associate, and Full Professor in Aeronautics and Astronautics at Stanford from 1998 to 2007, and in 2005 joined Berkeley. Claire works in the area of control theory and hybrid systems, with applications to air traffic management, UAV systems, energy, robotics, and systems biology.  She is a MacArthur Foundation Fellow (2006) and an IEEE Fellow (2010), and in 2010 held the Tage Erlander Professorship of the Swedish Research Council at KTH in Stockholm.
\end{IEEEbiography}

\end{document}